\def\year{2022}\relax
\documentclass[letterpaper]{article} % DO NOT CHANGE THIS
\usepackage{aaai22}  % DO NOT CHANGE THIS
\usepackage{times}  % DO NOT CHANGE THIS
\usepackage{helvet}  % DO NOT CHANGE THIS
\usepackage{courier}  % DO NOT CHANGE THIS
\usepackage[hyphens]{url}  % DO NOT CHANGE THIS
\usepackage{graphicx} % DO NOT CHANGE THIS
\usepackage{natbib}  % DO NOT CHANGE THIS AND DO NOT ADD ANY OPTIONS TO IT
\usepackage{caption} % DO NOT CHANGE THIS AND DO NOT ADD ANY OPTIONS TO IT
\DeclareCaptionStyle{ruled}{labelfont=normalfont,labelsep=colon,strut=off} % DO NOT CHANGE THIS
\usepackage{algorithm}
\usepackage{algorithmic}
\usepackage{amsmath}
\usepackage{amsfonts}
\usepackage{bbm}
\usepackage{booktabs}
\usepackage{multirow}
\usepackage{xcolor}
\usepackage{caption}
\usepackage{subcaption}
\definecolor{applegreen}{rgb}{0.0, 0.5, 0.0}
\usepackage{colortbl}
\usepackage{bbm}

\newcommand{\diag}{\mathop{\mathrm{diag}}}

\newtheorem{definition}{Definition} % for definition format
\newtheorem{theorem}{Theorem}
\newtheorem{lemma}[theorem]{Lemma} % for lemma format
\usepackage{tikz}
\newcommand*\circled[1]{\tikz[baseline=(char.base)]{
            \node[shape=circle,draw,inner sep=0.1pt, fill= yellow] (char) {#1};}} % for the symbol in flowchart

\usepackage{newfloat}
\usepackage{listings}
\floatstyle{ruled}
\newfloat{listing}{tb}{lst}{}
\floatname{listing}{Listing}
\title{BScNets: Block Simplicial Complex Neural Networks}
\author{
    Yuzhou Chen,$^1$
    Yulia R. Gel,$^2$
    H. Vincent Poor$^1$
}
\affiliations{
     \textsuperscript{\rm 1}Department of Electrical and Computer Engineering, Princeton University\\
     \textsuperscript{\rm 2}Department of Mathematical Sciences, University of Texas at Dallas\\
     \{yc0774, poor\}@princeton.edu, ygl@utdallas.edu\\
}

\usepackage{bibentry}
\begin{document}

\maketitle

\begin{abstract}
Simplicial neural networks (SNN) have recently emerged as the newest direction in graph learning which expands the idea
of convolutional architectures from node space to simplicial complexes on graphs. Instead of pre-dominantly assessing pairwise relations among nodes as in the current practice, 
simplicial complexes allow us to describe higher-order interactions and multi-node graph structures. By building upon connection between the convolution operation and the new block Hodge-Laplacian, we propose the first SNN for link prediction. Our new Block Simplicial Complex Neural Networks (BScNets) model generalizes the existing graph
convolutional network (GCN) frameworks by systematically incorporating salient interactions among multiple higher-order graph structures of different dimensions.
We discuss theoretical foundations behind BScNets and illustrate its utility for link prediction on eight real-world and synthetic datasets. Our experiments indicate that BScNets outperforms the state-of-the-art models by a significant margin 
while maintaining low computation costs. %, especially when integrated with the DropEdge technique. 
Finally, we show utility of BScNets as the new promising alternative for tracking spread of infectious diseases such as COVID-19 and measuring the effectiveness of the healthcare risk mitigation strategies. 
\end{abstract}

\section{Introduction}

Graph Convolutional Networks (GCNs) %and, more generally, tools of geometric deep learning (GDL), %i.e., deep learning architectures specifically designed for modeling data in non-Euclidean spaces such as graphs and manifolds.
is a powerful machinery for graph learning, allowing for efficient exploration of various pairwise interactions among graph nodes. However, most GCN-based approaches tend to be limited in their ability to efficiently exploit and propagate information across higher-order structures~\cite{morris2019weisfeiler, xiao2020graph}. In turn, many recent studies on cyber-physical, social, and financial networks suggest that relations among multi-node graph structures, as opposed to pairwise interaction among nodes, may be the key toward understanding hidden mechanisms behind structural organization of complex network systems. For example, disease transmission might be influenced not only by one-to-one interactions but also be largely impacted by various group relations and social reinforcement induced by each person’s social circle, e.g., COVID-19 anti-mask and anti-vaccine views. To enhance the obfuscation efforts in money laundering schemes, criminals involve complex interactions not only among multiple individuals but also among multiple criminal groups. Such higher-order graph interactions beyond the node space and multi-node structures
may be naturally modelled using simplicial complexes.

Simplicial neural networks (SNN) is the newly emerging direction 
in graph learning which extends convolutional operation to data that live on simplicial complexes. SNNs have recently been successfully applied to graph classification and trajectory forecasting~\cite{ebli2020simplicial, bunch2020simplicial, roddenberry2021principled, bodnar2021weisfeiler}. The key engine behind SNNs is the Hodge--de Rham theory that generalizes the standard graph Laplacian which describes node-to-node interactions via edges to Hodge-Laplacian which allows us to model diffusion from edges to edges via nodes, edges to edges via triangles, triangles to triangles via edges, etc~\cite{lim2015hodge, schaub2020random}. Furthermore, Hodge-Laplacian establishes a natural connection between
higher-order properties of discrete representations, e.g., graphs, and continuous representations, e.g., manifolds. As such, Hodge-Laplacian-based analytics opens multiple new perspectives for geometric representations at the interface of shape analysis, graph theory, and geometric deep learning (GDL)~\cite{Wang2019intrinsic, hajij2020cell}.  

 In this paper we make the first step toward bridging SNNs with link prediction on graphs. We propose a new Block Simplicial Complex Neural Networks (BScNets) model, by building upon the connection between convolution operation and block Hodge-style representation.
 In contrast to other SNNs which tend to focus only on edge-to-edge information flows, or Hodge 1-Laplacian,  BScNets allows us to simultaneously incorporate salient interactions among multiple simplicial complexes on graphs. Specifically, our BScNets scheme is composed of two components: the Adaptive Hodge Laplacian Based Block (AHLB) which describes multi-level structures and adaptively learns hidden dependencies among geometric representations of higher-order structures, and the Hodge-Style Adaptive Block Convolution (H-ABC) Module which automatically infers relations among multi-dimensional simplices. Our results indicate that this novel integration of information flows across not one but multiple higher-order structures via the Block Hodge Laplacian analytics yields the highest performance in link prediction tasks. 
 %We discuss theoretical properties of the random walk-based block Hodge-Laplacian (i.e., the key engine behind BScNets) and extensively evaluate BScNets performance for link prediction on real-world and synthetic datasets, from such diverse domains as criminal, social, and transportation networks. %Furthermore, we propose an adaptive Hodge Laplacian based block operator which enables us to simultaneously integrate knowledge on interactions among higher order substructures of various orders into graph learning.%adaptive Hodge Laplacian based block operator

Significance and novelty of our contributions can be summarized as follows:
\begin{itemize}
    \item Our BScNets is the first SNN for link prediction on graphs, bridging the recently emerging concepts of convolutional architectures on simplicial complexes with topological signal processing on graphs.
    
    \item We propose a new (random-walk based) adaptive Hodge Laplacian based block operator which simultaneously integrates knowledge on interactions among multiple higher-order graph structures and discuss its theoretical properties.
    
    \item We extensively validate BScNets on real-world and synthetic datasets, from such diverse domains as criminal, collaboration and transportation networks. Our results indicate that BScNets outperforms the state-of-the-art models by a significant margin while maintaining low computation costs.
    
    \item We discuss utility of BScNets and SNN tools as the new promising alternative for tracking spread of infectious diseases such as COVID-19 and evaluating healthcare risk mitigation strategies. 
    
\end{itemize}

\section{Related Work}

{\bf Link Prediction} GCN-based methods are known to be the powerful machinery for link prediction tasks. Specifically, the Graph Autoencoder (GAE)~\cite{kipf2016variational} and its variational version, i.e., Variational Graph Autoencoder (VGAE), are first employed to link prediction on citation networks. SEAL~\cite{zhang2018link} extracts local enclosing subgraphs around the target links and learns a function mapping the subgraph patterns to link existence. %Graph2Gauss (G2G)~\cite{bojchevski2018deep} designs an unsupervised model that handles inductive link prediction by using deep encoder to embed each node as a Gaussian distribution. 
In addition, the Hyperbolic Graph Convolutional Neural Networks (HGCN)~\cite{chami2019hyperbolic} leverages both the hyperbolic geometry and GCN framework to learn node representations. Another interesting recent strategy is to use pairwise topological features to find latent representations of geometrical structure of graph using GCN~\cite{yan2021link}. Our method differs from these approaches in that it explicitly models the higher-dimensional graph substructures and higher-order interactions via building an adaptive and interpretable Hodge block representation. Moreover, we propose a novel Hodge-style adaptive block convolution module to aggregate topological features encoded in the simplicial complexes by investigating relationships between simplices of different orders. This higher-order simplicial representation is substantially more general than the structural representation of node sets.

{\bf Simplicial Neural Networks} Modeling higher-order interactions on graphs is an emerging direction in graph representation learning. 
While the role of higher-order graph structures for graph learning has been documented for a number of years~\cite{agarwal2006higher, johnson2012discrete} and involve such diverse applications as graph signal processing in electric, transportation and neuroscience systems, 
including link prediction~\cite{benson2018simplicial}, integration of higher-order graph substructures into deep learning on graphs has emerged only in 2020.
%As \citet{benson2018simplicial, schaub2020random} showed, higher-order network structures can be leveraged to boost the performance of link and trajectory prediction tasks.
Indeed, several most recent studies %~\cite{ebli2020simplicial, bunch2020simplicial, roddenberry2021principled, bodnar2021weisfeiler} 
propose to leverage simplicial information to perform neural networks on graphs. For instance,~\citet{ebli2020simplicial} develops a model called Simplicial Neural Networks, which integrates the simplicial Laplacian of a simplicial complex into neural network framework based on simplicial Laplacian. Message Passing Simplicial Networks (MPSNs)~\cite{bodnar2021weisfeiler} is proposed by performing massage passing on simplicial complexes for graph classification. Similarly, SCoNe~\cite{roddenberry2021principled} uses the GCN architecture depended on simplicial complexes to extrapolate trajectories on trajectory data. Besides, the convolutional message passing scheme on cell complex~\cite{hajij2020cell} is shown to facilitate representational learning on graphs. Our approach further advances these recent results
by explicitly exploiting local topological information encoded in simplicial complexes of multiple dimensions and extracting the key interaction relations among multiple higher-order graph structures, which leads to significant gains in link prediction accuracy. 

\section{Block Simplicial Complex Neural Networks}

We consider a graph $\mathcal{G} = \left(\mathcal{V}, \mathcal{E}, \boldsymbol{X}\right)$, where $\mathcal{V}$ is the set of nodes ($|\mathcal{V}| = n$) and $\mathcal{E} \subseteq \mathcal{V} \times \mathcal{V}$ is the set of edges ($|\mathcal{E}| = m$). Furthermore, each node $v_i$ is associated with a $d$-dimensional vector of attributes $\boldsymbol{x}_i$ which is the $i$-th row of matrix $\boldsymbol{X} \in \mathbb{R}^{n \times d}$ (where $d$ is the input feature dimension). Connectivity of $\mathcal{G}$ can be encoded in a form of an adjacency matrix $\boldsymbol{A} \in \mathbb{R}^{n \times n}$ with entries $[\boldsymbol{A}]_{ij}=1$ if nodes $i$ and $j$ are connected and 0, otherwise. For undirected graph $\mathcal{G}$, $\boldsymbol{A}$ is symmetric (i.e.,  $\boldsymbol{A}=\boldsymbol{A}^{\top}$).

%Given the union set of positive (observed) and negative (unobserved) training links $\mathcal{E}^{\text{Train}}_p \cup \mathcal{E}^{\text{Train}}_n$, and node features $\boldsymbol{X}$, we intend to obtain a model $\mathfrak{F}$ to predict the existence of the link between nodes $u_i$ and $u_j$ (where $u_i, u_j \in \mathcal{V}$) in the union set of positive and negative testing links $\mathcal{E}^{\text{Test}}_p \cup \mathcal{E}^{\text{Test}}_n$. This can be defined as $(u_i, ?, u_j) = \mathfrak{F}(\mathcal{E}^{\text{Train}}_p \cup \mathcal{E}^{\text{Train}}_n, \boldsymbol{X})$.

{\bf Background on %Simplicial Complexes and 
Hodge Theory} One of the focal points of graph theory and, in virtue of it, GDL, is graph Laplacian. Laplacian allows us to establish a natural link between discrete representations, e.g., graphs, and continuous representations, e.g.,  manifolds~\cite{chung1997spectral}.
%(Here $\boldsymbol{D}$ is the matrix of degrees, i.e., $\boldsymbol{D}_{ii}=\sum_{j}\boldsymbol{A}_{ij}$. In this paper, we primarily consider a normalized version of graph Laplacian $\boldsymbol{L} = \boldsymbol{I} - \boldsymbol{D}^{-1/2}\boldsymbol{A}\boldsymbol{D}^{-1/2}$, where $\boldsymbol{I}$ is the identity matrix.) 
The (unnormalized) graph Laplacian is defined as $\boldsymbol{L}_0 = \boldsymbol{D}-\boldsymbol{A}$, and $\boldsymbol{L}_0$ is a symmetric and positive-semidefinite matrix. The Laplacian $\boldsymbol{L}_0$ represents a discrete counterpart of the Laplacian operator in vector calculus. In particular, as divergence of the gradient of some twice-differentiable multivariate function in vector calculus, the Laplacian operator is the flux density of the gradient flow of this function. That is, Laplacian measures how much the value of the function at any given point differs from average values of the function evaluated at nearby points. In turn, in the discrete case similarly, graph Laplacian $\boldsymbol{L}_0$ measures diffusion from node to node through edges and, broadly speaking, assesses at what extent the graph $\mathcal{G}$ differs at one node from the graph $\mathcal{G}$ at surrounding nodes. While $\boldsymbol{L}_0$ contains some very important information on the topology of $\mathcal{G}$, the natural question arises {\it what if we are interested in diffusion dynamics on graph substructures beyond the node level?}
For example, formation of money laundering activities within criminal networks by default involves very complex multi-node interactions and, hence, cannot be well captured by methods that focus on the dyadic graph relationships.  To assess such higher-order network properties, we can study graphs in terms of topological objects such as {\it simplicial complexes} and exploit the discrete Hodge theory, particularly, Hodge-Laplacian-based analytics as generalization of Laplacian dynamics to polyadic substructures of $\mathcal{G}$.

\begin{definition}
A family $\Delta$ of finite subsets of a set $\mathcal{V}$ is
an {\it abstract simplicial complex} if for every $\sigma\in\Delta$,  $\tau\subseteq\sigma$ implies $\tau\in\Delta$. I.e., $\Delta$ is closed under the operation of taking subsets. If $|\sigma|=k+1$, then $\sigma$ is called a \emph{$k$-simplex}.
Every subset $\tau\subset\sigma$ such that $|\sigma|=k$ is called a {\it face} of $\sigma$. All simplices in $\Delta$ that have $\sigma$ as face are called {\it co-faces}. Dimension of $\Delta$ is the largest dimension of any of its faces, or
$\infty$ if there is no upper bound on the dimension of the faces.
\end{definition}

Hence, %viewing $\mathcal{G}$ as a 2-dimensional simplicial complex formed by $\mathcal{V}$ and $\mathcal{E}$,  
nodes of $\mathcal{G}$ are 0-simplices, edges are 1-simplices, and triangles are 2-simplices. For a $k$-simplex of $k>0$, we can also define its {\it orientation} by
(arbitrary) selecting some order for its nodes, and two orderings are said to be equivalent if they differ by an even permutation. As a result,
for a given $k$-simplex $\sigma$ with orientation  $[i_0, i_2,\ldots, i_k]$, any face of $\sigma$ is assigned its own orientation (or ``identifyer") $[i_0, i_1,\ldots, i_{j-1}, i_{j+1}, \ldots,  i_k]$ (i.e., we omit the $j$-th element). 
To study diffusion among higher-order substructures of $\mathcal{G}$, we now form a real-valued vector space $C^k$ which is endowed with basis from the oriented $k$-simplices and whose elements are called {\it $k$-chains}. Diffusion through higher-order graph substructures can be then defined via linear maps among spaces $C^k$ of $k$-chains on $\mathcal{G}$~\cite{lim2015hodge}.

\begin{definition}
A linear map $\partial_k: C^k \rightarrow C^{k-1}$ is called a {\it boundary} operator. The adjoint of the boundary map induces the {\it co-boundary} operator
$\partial_k^{T}: C^k \rightarrow C^{k+1}$.
Matrix representations of $\partial_k$
and $\partial_k^{\top}$ are $\boldsymbol{B}_k$ and $\boldsymbol{B}_k^{\top}$, respectively.

An operator over oriented $k$-simplices $\boldsymbol{L}_k:  C^k \rightarrow C^k$ is called the {\it $k$-Hodge Laplacian}, and its matrix representation is given by
\begin{eqnarray}
\label{Hodge}
\boldsymbol{L}_k=\boldsymbol{B}_k^{\top}\boldsymbol{B}_k+\boldsymbol{B}_{k+1}\boldsymbol{B}_{k+1}^{\top},\end{eqnarray}
where $\boldsymbol{B}_k^{\top}\boldsymbol{B}_k$ and
$\boldsymbol{B}_{k+1}\boldsymbol{B}_{k+1}^{\top}$ 
are often referred to $\boldsymbol{L}^{down}_k$ and $\boldsymbol{L}^{up}_k$, respectively.
\end{definition}
As $\boldsymbol{B}_0 = 0$, the standard graph Laplacian $\boldsymbol{L}_0$ is a subcase of~(\ref{Hodge}) which tracks diffusion process from nodes to nodes via edges. Indeed, $\boldsymbol{L}_0 = \boldsymbol{B}_1\boldsymbol{B}_1^{\top}$, where $\boldsymbol{B}_1$ is an $n\times m$-incidence matrix of $\mathcal{G}$  (i.e., $\boldsymbol{B}_{1}[ij]=1$ if node $i$ and edge $j$ are incident and 0, otherwise).
In turn, $\boldsymbol{L}^{down}_1=\boldsymbol{B}_1^{\top}\boldsymbol{B}_1$ is often referred to as {\it edge Laplacian} and assesses diffusion from edges to edges via nodes. Finally, Hodge 1-Laplacian $\boldsymbol{L}_1$ 
measures variation on functions defined on graph edges (i.e., 1-simplices) with respect to the incidence relations among edges and nodes (i.e., 0-simplices) and edges and triangles (i.e., 2-simplices). More generally, $\boldsymbol{L}_k$ measures variation on functions defined on $k$-simplices of $\mathcal{G}$ with respect to incidence relations among $k$-simplices with $(k-1)$- and $(k+1)$-simplices. Figure~\ref{B1_B2_illustration} shows an example of the simplicial complex on a graph (more details are in Appendix A.1). 
\begin{figure}[h!]
    \centering
    \includegraphics[width=0.4\textwidth]{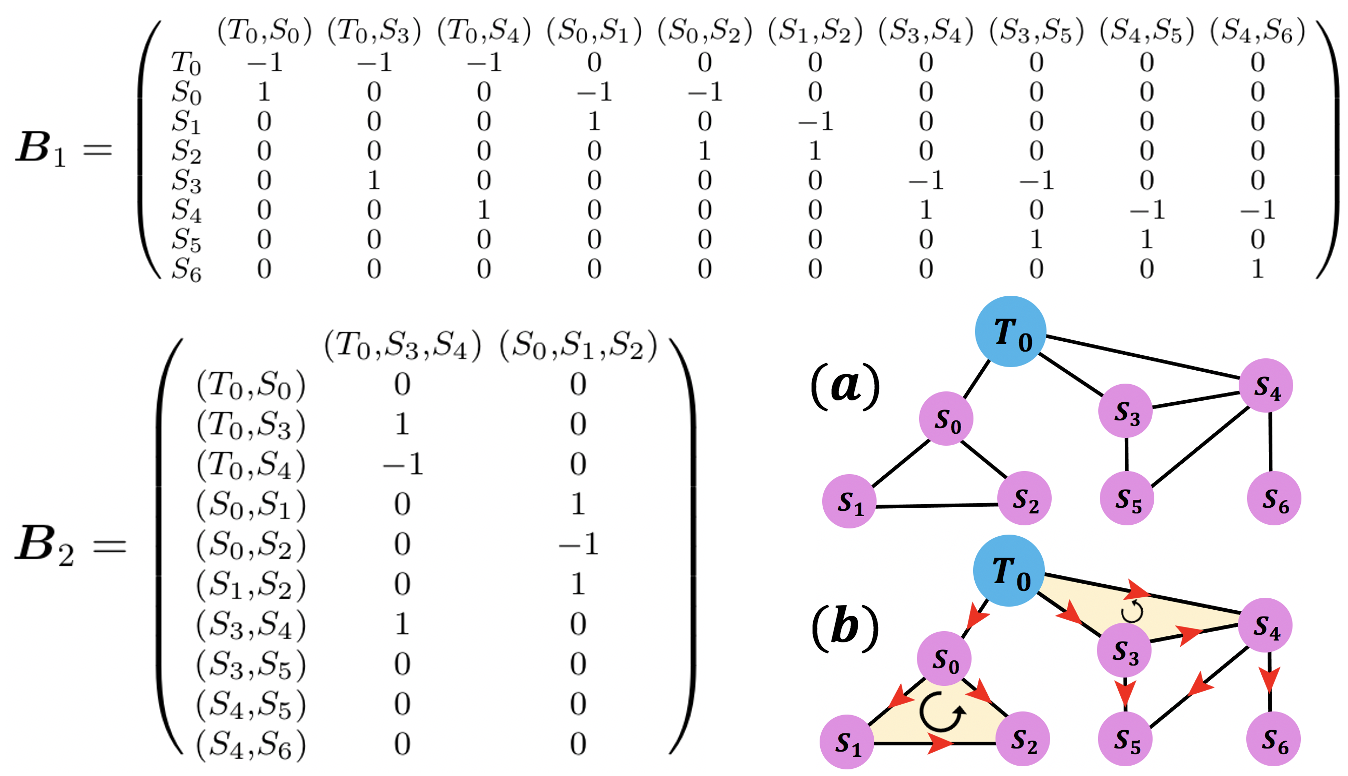}
    \caption{An example of generating simplicial complex from a classroom environment network, where $T_0$ is the teacher\_0 and $S_u$ is the student\_$u$ ($u \in \{0, \dots, 6\}$). (a) Graph structure of the classroom environment network; (b) Simplicial complex of the classroom environment network; $\boldsymbol{B}_1$ and $\boldsymbol{B}_2$ are the corresponding node-to-edge and edge-to-face incidence matrices, respectively.}
    \label{B1_B2_illustration}
\end{figure}

%\YC{Should we give an example for $B_1$ and $B_2$?}
%\YGL{Yes, we shall as otherwise it's hard to get the idea. We can refer to the examples of $B_1$ and $B_2$ from the background section too.}

{\bf Random Walk Based Block Hodge-Laplacian Operator} While the Hodge theory allows us to systematically describe diffusion across higher-order graph substructures, or $k$-simplices of any $k$, all current studies are restricted solely to Hodge 1-Laplacian $\boldsymbol{L}_1$~\cite{schaub2020random, roddenberry2021principled, bodnar2021weisfeiler}.
In this paper we propose a new random walk based block Hodge-Laplacian operator which enables us to simultaneously integrate knowledge on interactions among higher-order substructures of various orders into graph learning.

\begin{definition} The $K$-block Hodge-Laplacian $\boldsymbol{\mathfrak{L}}_K$ is a divergence operator on
the direct sum of vector spaces $C^k$, $k\in \mathbb{Z}_{\geq 0}$. That is, $\boldsymbol{\mathfrak{L}}_K:\oplus_{k=0}^K C^k \rightarrow \oplus_{k=0}^K C^k$.
Block Hodge-Laplacian $\boldsymbol{\mathfrak{L}}_K$ can be represented as a diagonal block matrix $\diag\{\boldsymbol{L}_0, \boldsymbol{L}_1, \ldots, \boldsymbol{L}_K\}$.
Furthermore, by selecting a subset of indices $\{k_1, \ldots, k_j, \ldots, k_J\}\in \mathbb{Z}_{\geq 0}$, we can consider a reduced
block Hodge-Laplacian $^R\boldsymbol{\mathfrak{L}}_J$, 
with a matrix representation
 $\diag\{\boldsymbol{L}_{k_1}, \boldsymbol{L}_{k_2}, \ldots, \boldsymbol{L}_{k_J}\}$, which allows us to describe interaction relations among a particular subset of higher-order graph substructures. Here positive integers $K$ and $k_J$ are bounded by the dimension of the abstract simplicial complex $\Delta$ on $\mathcal{G}$
and in practice, $K$ and $k_J$ are typically $\leq 2$.
\end{definition}

The $K$-block Hodge-Laplacian $\boldsymbol{\mathfrak{L}}_K$ is related to the Dirac operator in differential geometry (i.e., Dirac operator is a square root of $\boldsymbol{\mathfrak{L}}_K$)~\cite{lloyd2016quantum}. As such, $\boldsymbol{\mathfrak{L}}_K$ has multiple implications for analysis of synchronization dynamics and coupling of various topological signals on graphs, with applications in physics and quantum information processing~\cite{calmon2021topological}.

\begin{lemma}
For $K, J\in \mathbb{Z}_{\geq 0}$, operators $\boldsymbol{\mathfrak{L}}_K$ and $^R\boldsymbol{\mathfrak{L}}_J$ are symmetric and semipositive-definite, i.e. $\boldsymbol{\mathfrak{L}}_K\geq 0$ and  $^R\boldsymbol{\mathfrak{L}}_J \geq 0$;
 $\boldsymbol{\mathfrak{L}}_K=(\boldsymbol{\mathfrak{L}}_{K})^{\top}$
and $^R\boldsymbol{\mathfrak{L}}_J=(^R\boldsymbol{\mathfrak{L}}_J)^{\top}$.
\end{lemma}

Furthermore, we propose the {random walk-based}  block Hodge-Laplacian, i.e., $r$-{th} power of block Hodge-Laplacian representation (where $r \in \mathbb{Z}_{>0}$). Our {random walk-based} block Hodge-Laplacian is inspired by the recent success in random walk based graph embeddings and simplicial complexes %~\cite{grover2016node2vec, abu2019mixhop, benson2018simplicial, bodnar2021weisfeiler} 
but is designed to conduct informative {\it joint} random walks on higher-order Hodge Laplacians instead of limited powering Hodge 1-Laplacian~\cite{benson2018simplicial, bodnar2021weisfeiler}. 
Indeed, successfully travelling through higher-order topological features will provide us with additional  feature information which is valuable for learning edge embeddings. 
The following Lemma~\ref{lemma_power} provides insight on the interplay among random walks on various $k$-simplices and their respective Hodge-Laplacian representation. 
%Random walks on Hodge Laplacian are stochastic processes that successively travels among higher-order topological features (e.g., nodes, edges, and triangles) which boost the deep higher-order representation learning and take higher-order interaction order information into consideration. 

\begin{lemma}\label{lemma_power}
For any $r\in \mathbb{Z}_{>0}$ and $K, J\in \mathbb{Z}_{\geq 0}$, the $r$-power of 
$(\boldsymbol{\mathfrak{L}}_K)^r$ and $(^R\boldsymbol{\mathfrak{L}}_J)^r$
are given by 
$\oplus_{k=0}^K ((\boldsymbol{L}^{down}_k)^r+(\boldsymbol{L}^{up}_k)^r)$
and $\oplus_{j=1}^J ((\boldsymbol{L}^{down}_{k_j})^r+(\boldsymbol{L}^{up}_{k_j})^r)$, respectively.
\end{lemma}

{\bf Adaptive Hodge Laplacian Based Block Analytics}
Our new approach to learning higher-order graph topology is motivated by the following question: 
{\it Can we capture the interaction relations among $k$-simplices on the graph $\mathcal{G}$ whose orders are farther than $k-1$ and $k+1$ apart?}
While addressing this problem falls outside the Hodge-de Rham theory on the simplicial
Hodge Laplacians, we provide an affirmative answer to this question through building an adaptive Hodge Laplacian Based Block Operator.
%{\it To summarize the node information into higher-order graph representation, can message passing mechanism be incorporated into higher-dimensional topological space (e.g., simplicial complex) and how to capture the relationship between different dimensions of the complex?} 

Without loss of generality, let us consider a 2-block (reduced) Hodge-Laplacian 
\begin{eqnarray}
\label{2HL}
\boldsymbol{\mathfrak{L}}_2=\left[
\begin{array}{c|c}
\boldsymbol{L}_{k_1} & O \\ \hline
O & \boldsymbol{L}_{k_2}
\end{array}\right].
\end{eqnarray}
Our goal is to construct a new linear operator such that we describe interaction relations among $\boldsymbol{L}_{k_1}$ and $\boldsymbol{L}_{k_2}$. This problem can be addressed by defining and learning a similarity function $f(\boldsymbol{L}_{k_1}, \boldsymbol{L}_{k_2})$. The natural choice for such similarity function is the inner product 
among elements in $C^{k_1}$ (i.e., $k_1$-chains) and  elements in $C^{k_2}$ (i.e., $k_2$-chains). However, in general, $dim (C^{k_1})\neq dim (C^{k_2})$. Assume
that $d_1=dim (C^{k_1})> d_2=dim (C^{k_2})$. Hence, instead we can consider a similarity function based on the inner product
between elements in $C^{k_2}$ and elements in $\mathbb{P}_{d_{2}}C^{k_1}$, i.e., the projection of $C^{k_1}$ on the lower dimensional space. Here $\mathbb{P}_{d_2}$ is the corresponding orthogonal projector and, given symmetry of $\boldsymbol{L}_{k_1}$, can be formed by the eigenvectors of $\boldsymbol{L}_{k_1}$ corresponding to the top $d_2$ largest eigenvalues.  
Our new Adaptive Hodge Laplacian Based (AHLB) Block operator then takes the form
%$$
\begin{align}
\label{ahlb_laplacian}
\boldsymbol{\mathbbm{L}}^{B}_2=\left[
\begin{array}{c|c}
\boldsymbol{L}_{k_1} & f(\mathbb{P}_{d_{2}}\boldsymbol{L}_{k_1}, \boldsymbol{L}_{k_2}) \\ \hline
f(\mathbb{P}_{d_{2}}\boldsymbol{L}_{k_1}, \boldsymbol{L}_{k_2})^{\top} & \boldsymbol{L}_{k_2}
\end{array}\right].
\end{align}
%$$
Note that in general, linear operator $\boldsymbol{\mathbbm{L}}^B_2 \in \mathbb{R}^{(d_1 + d_2) \times (d_1 + d_2)}$ is no longer a Laplacian. For example, while $\boldsymbol{\mathbbm{L}}^B_2$ is symmetric (by construction), it may not satisfy the condition of positive semidefinitess (see Lemma 3 in Appendix A.2). However,
as shown below, AHLB opens multiple opportunities to better describe higher-order interactions on graphs that are inaccessible not only with individual $k$-Hodge Laplacians but even with the $K$-block Hodge-Laplacian operator (see the ablation study for more details). In addition, followed by Lemma~\ref{lemma_power}, we also consider a case when AHLB is constructed based on the $r$-th power of $\boldsymbol{\mathfrak{L}}_2$, and our studies indicate that on average the best results are achieved for $r=2$ (see Table~\ref{ablation_architecture} and Appendix A.3).

Armed with the AHLB operator (Eq.~\ref{ahlb_laplacian}), we can utilize the non-local message operation $f(\cdot,\cdot)$~\cite{wang2018non} to capture long-range relations and intrinsic higher-order connectivity among entities in the 2-block (reduced) Hodge-Laplacian $\boldsymbol{\mathfrak{L}}_2$, i.e., $f(\mathbb{P}_{d_{2}}\boldsymbol{L}_{k_1}, \boldsymbol{L}_{k_2})$ placed in the off-diagonal. We now describe the choices of non-local message passing functions which can be used for $f(\cdot,\cdot)$ in the following. Specifically, given two higher-order Hodge Laplacians $\boldsymbol{L}_{k_1}$ and $\boldsymbol{L}_{k_2}$, we define two types of the non-local message passing functions to capture the relations between $[\mathbb{P}_{d_{2}}\boldsymbol{L}_{k_1}]_i \in \mathbb{R}^{d_2}$ and $[\boldsymbol{L}_{k_2}]_j \in \mathbb{R}^{d_2}$ (i.e., topological embedding of the $i$-th graph substructure in $\mathbb{P}_{d_{2}}\boldsymbol{L}_{k_1}$ and topological embedding of the $j$-th graph substructure in $\boldsymbol{L}_{k_2}$ respectively) as

{(1) \textit{Inner-Product}}
$$f([\mathbb{P}_{d_{2}}\boldsymbol{L}_{k_1}]_i, [\boldsymbol{L}_{k_2}]_j) =
<[\mathbb{P}_{d_{2}}\boldsymbol{L}_{k_1}]_i, [\boldsymbol{L}_{k_2}]_j>.
%= ([\mathbb{P}_{d_{2}}\boldsymbol{L}_{k_1}]_i)^{\top} \cdot [\boldsymbol{L}_{k_2}]_j,
$$
Since all considered Laplacians in our case are over the real field $\mathbb{R}$, we can consider a dot-product. However, in a more general case of multivariable functions on graph simplices, $f(\cdot, \cdot)$ can be an inner-product.  

{(2) \textit{Embedded Inner-Product}} %$f([\mathbb{P}_{d_{2}}\boldsymbol{L}_{k_1}]_i, [\boldsymbol{L}_{k_2}]_j)$
$$ f([\mathbb{P}_{d_{2}}\boldsymbol{L}_{k_1}]_i, [\boldsymbol{L}_{k_2}]_j) =
<\boldsymbol{\Theta}_{\xi} [\mathbb{P}_{d_{2}}\boldsymbol{L}_{k_1}]_i, \boldsymbol{\Theta}_{\psi}[\boldsymbol{L}_{k_2}]_j>,
%(\boldsymbol{\Theta}_{\xi} [\mathbb{P}_{d_{2}}\boldsymbol{L}_{k_1}]_i)^{\top} \cdot \left(\boldsymbol{\Theta}_{\psi}[\boldsymbol{L}_{k_2}]_j\right),
$$
where $\boldsymbol{\Theta}_{\xi} \in \mathbb{R}^{d_c \times d_2}$ and $\boldsymbol{\Theta}_{\psi} \in \mathbb{R}^{d_c \times d_2}$ are weight matrices to be learned, and $d_c$ is the embedding dimension. As such, we infer the relation value between two simplices.

%Therefore, to compute relation value between two entities in $\boldsymbol{L}_{k_1}, \boldsymbol{L}_{k_2}$, we can use either {\it Inner-Product} or {\it Embedded Inner-Product} as the relation function for $f(\cdot,\cdot)$. 
Finally, the adaptive Hodge block can be formulated as
\begin{align}
\tilde{\boldsymbol{\mathbbm{L}}}^B_2 = \text{softmax}
\left(\text{ReLU}\left(\boldsymbol{\mathbbm{L}}^B_2\right)\right),
\end{align}
where $\text{softmax}$ function is applied to normalize the block operator $\boldsymbol{\mathbbm{L}}^B_2$, and the $\text{ReLU}$ activation function (where $\text{ReLU}(\cdot) = \max{(0, \cdot)}$) eliminates both weak pairwise relation among higher-dimensional simplices and weak connections. As a result, the adaptive Hodge block $\tilde{\boldsymbol{\mathbbm{L}}}^B_2$ can dynamically learns interactions among simplices of different dimensions.

\medskip
{\bf Hodge-Style Adaptive Block Convolution Module}
Armed with the AHLB operator, we now discuss how to construct the Hodge-style adaptive block convolution and use it to learn the distance between two nodes in the embedding space. Instead of using graph convolutional operator, i.e., $\boldsymbol{L}_0$, we adopt the AHLB operator $\tilde{\boldsymbol{\mathbbm{L}}}^B_2$ to demonstrate the effectiveness of the proposed novel higher-order Hodge convolution operator. The Hodge-style adaptive block convolution (H-ABC) module is given by
\begin{align}
\label{h_abc_eq}
    {\boldsymbol{Z}^{(\ell + 1)}} = \left(\tilde{\boldsymbol{\mathbbm{L}}}^B_2{\boldsymbol{Z}^{(\ell)}}{\boldsymbol{\Theta}^{(\ell)}_1} \right){\boldsymbol{\Theta}^{(\ell)}_2},
\end{align}
where $\boldsymbol{Z}^{(0)} = \boldsymbol{X} \in \mathbb{R}^{n \times d}$ is node features matrix, $\boldsymbol{\Theta}^{(\ell)}_1 \in \mathbb{R}^{d \times (d_1 + d_2)}$ and $\boldsymbol{\Theta}^{(\ell)}_2 \in \mathbb{R}^{(d_1 + d_2) \times d\textsubscript{out}}$ are two trainable weight matrices at layer $\ell$, and $d\textsubscript{out}$ denotes the dimension of node embedding at the $(\ell)$-th layer through the H-ABC operation. For the link prediction task, we use the Fermi-Dirac decoder~\cite{nickel2017poincare} to compute the distance between the two nodes. Formally,
\begin{align}
\label{h_abc}
    \text{dist}^{\text{H-ABC}}_{uv} = ({\boldsymbol{Z}^{(\ell + 1)}_u - \boldsymbol{Z}^{(\ell + 1)}_v})^2,
\end{align}
where $\text{dist}\textsuperscript{H-ABC}_{uv} \in \mathbb{R}^{1\times d\textsubscript{out}}$ is the distance between nodes $u$ and $v$ in a local spatial domain.

\medskip
{\bf Graph Convolution Layer} Similar to the process of computing distances between learnable node embeddings with the Hodge-style adaptive block convolution, we also use graph convolution operation (Eq.~\ref{gcn_rep}) to evaluate the distance (Eq.~\ref{gcn_distance}) between the node embeddings of nodes $u$ and $v$ as%, i.e., via geographic information equipped with node attributes.
\begin{align}
\label{gcn_rep}
    {\boldsymbol{H}^{(\ell +1)}} & = \tilde{\boldsymbol{L}} \boldsymbol{H}^{(\ell)}{\boldsymbol{\Theta}^{(\ell)}_3}, \\
    \text{dist}^{\text{GC}}_{uv} &= ({\boldsymbol{H}^{(\ell + 1)}_u - \boldsymbol{H}^{(\ell + 1)}_v})^2,\label{gcn_distance}
\end{align}
where $\tilde{\boldsymbol{L}} = {\boldsymbol{D}}_v^{-1/2}({\boldsymbol{A} + \boldsymbol{I}}){\boldsymbol{D}}_v^{-1/2}$ (where ${\boldsymbol{D}}_{v}$ is the degree matrix of $\boldsymbol{A} + \boldsymbol{I}$, i.e., $[\boldsymbol{D}_v]_{ii} = \sum_j [\boldsymbol{A} + \boldsymbol{I}]_{ij}$), $\boldsymbol{H}^{(0)} = {\bf X} \in \mathbb{R}^{n \times d}$ is node features matrix, ${\boldsymbol{\Theta}^{(\ell)}_3} \in \mathbb{R}^{d \times d\textsubscript{out'}}$ is the trainable weight matrix, $d\textsubscript{out'}$ denotes the dimension of node embedding at the $(\ell)$-th layer through the graph convolution operation, $\text{dist}^{\text{GC}}_{uv} \in \mathbb{R}^{1\times d\textsubscript{out'}}$ is the distance between nodes $u$ and $v$ in global spatial domain.

\begin{figure}[h!] % [t!]
    \centering
    \includegraphics[width=0.47\textwidth]{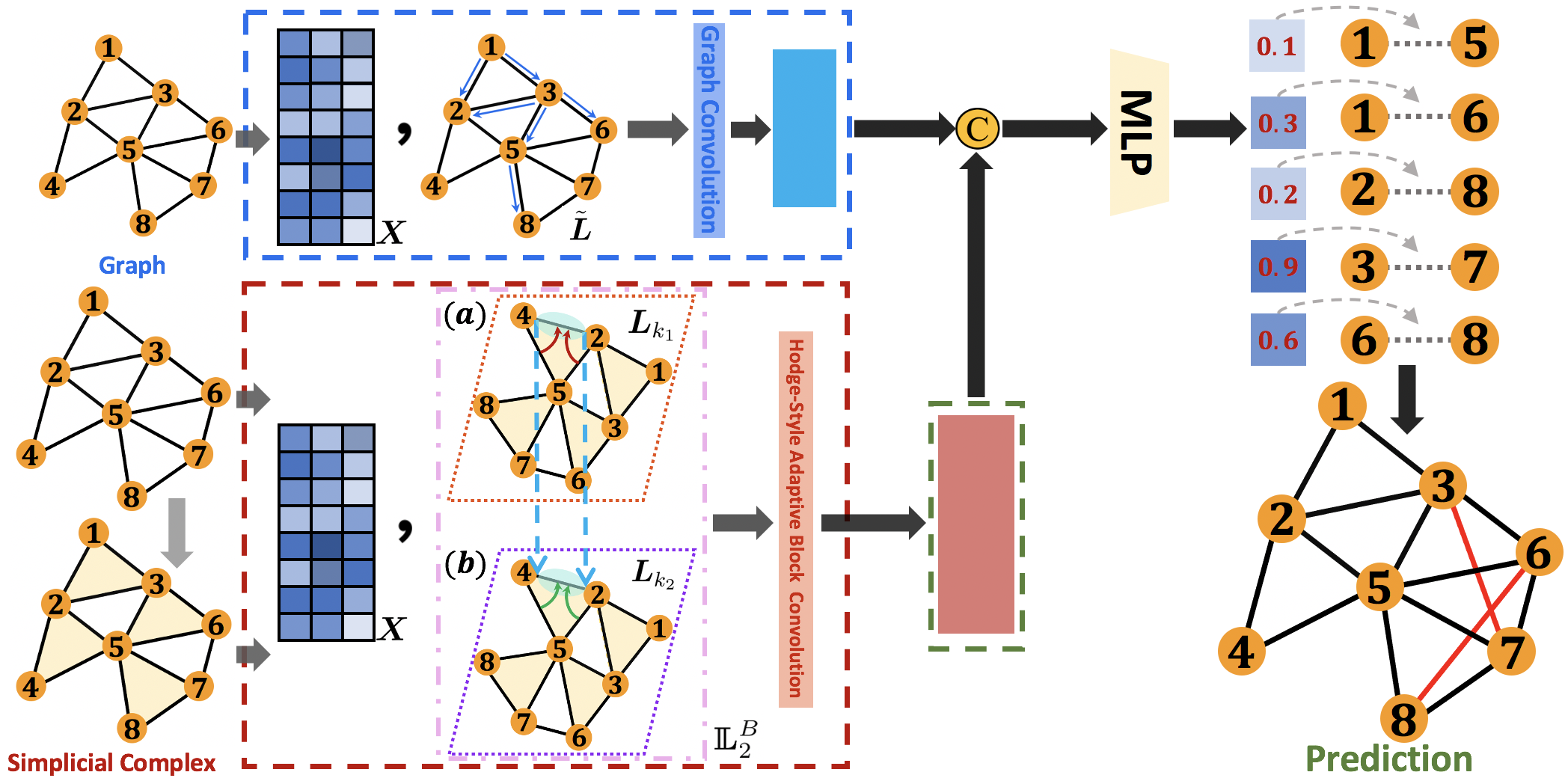}
    \caption{The architecture of BScNets. BScNets consists of graph convolution (upper row) and Hodge-style adaptive block convolution module (lower row). Each convolution layer/module contains the node feature matrix $\boldsymbol{X}$ and a Laplacian/operator (i.e., normalized Laplacian $\tilde{\boldsymbol{L}}$ and adaptive Hodge Laplacian based operator $\boldsymbol{\mathbbm{L}}^{B}_2$ for upper and lower parts respectively). Lower row presents the relation modeling between (a) Hodge Laplacian $\boldsymbol{L}_{k_1}$ describing diffusion from edge to edge through nodes and (b) Hodge Laplacian $\boldsymbol{L}_{k_2}$ describing diffusion from edge to edge through triangles. Symbol $\circled{C}$ represents concatenation.\label{flowchart}}
\end{figure}

\medskip
{\bf Distance between Two Nodes} We wrap the concatenation of Hodge-style adaptive convolution and graph convolution operation outputs (i.e., in Eqs.~\ref{h_abc} and~\ref{gcn_distance}) into Multi-layer Perceptron (MLP) block
\begin{align*}
    \text{dist} = \text{ReLU}(f_{\text{MLP}}([\pi_{\alpha}\times\text{dist}^{\text{GC}}, \pi_{\beta}\times\text{dist}^{\text{H-ABC}}])),
\end{align*}
where $[\cdot,\cdot]$ denotes the concatenation of the outputs from H-ABC module and graph convolution operation, $f_{\text{MLP}}$ is an MLP layer that maps the concatenated embedding to a $d_o$-dimensional space, and $\pi_{\alpha}$ and $\pi_{\beta}$ are the hyperparameters representing the weight of each distance (i.e., dist) in the $(\ell+1)$-th layer. Based on the propagation rule above, the edges connection probability can be computed as
%\begin{align}
    $prob(u,v)  = \left[{\exp{((\text{dist}_{uv}- \delta)/\eta)}+1}\right]^{-1}$,
%\end{align}
where $\delta$ and $\eta$ are hyperparameters. Then, training via standard backpropagation is performed via binary cross-entropy loss function using negative sampling. Figure~\ref{flowchart} illustrates our proposed BScNets framework, which consists of the Hodge-style adaptive block convolution module and graph convolution operation (see the Appendix A.4 for details). %Figure~\ref{flowchart} shows that in BScNets  (i) the graph convolution is employed directly on graph-structured data to model node embeddings by aggregating the embeddings of neighbors and (ii) the Hodge-style adaptive block convolution module is constructed to jointly model edge embeddings via (1) message passing from edges to edges through nodes, (2) massage passing from edges to edges through triangles, and (3) relation modeling among Hodge representations from two simplicies. Lastly, we concatenate two corresponding embeddings and deploy a MLP to predict whether an edge exists between any pair of nodes. % will put it in Appendix

%\begin{align*}
%    Loss_p(q) = -\frac{1}{N}\sum_{i=1}^N y_i \log{(p(y_i))} + (1-y_i) \log{(1-p(y_i))}
%\end{align*}

\section{Experimental Study}

{\bf Datasets} We experiment on three types of networks (i) {citation networks}: Cora and PubMed~\cite{sen2008collective}; (ii) social networks: (1) {flight network}: Airport~\cite{chami2019hyperbolic}, (2) {criminal networks}: Meetings and Phone Calls~\cite{cavallaro2020disrupting}, and (3) {contact networks}: High School network and Staff Community~\cite{salathe2010high,eletreby2020effects}; (iii) {disease propagation tree}: Disease~\cite{chami2019hyperbolic}. The statistics and more details of the datasets are provided in Appendix B.1.1.

{\bf Baselines} We compare against ten state-of-the-art (SOA) baselines, including (i) MLP, (ii) GCN~\cite{kipf2016semi}, (iii) Simplified Graph Convolution (SGC)~\cite{wu2019simplifying}, (iv) Graph Attention Networks (GAT)~\cite{velivckovic2018graph}, (v) GraphSAG (SAGE), (vi) SEAL~\cite{zhang2018link}, (vii) Hyperbolic neural networks (HNN)~\cite{ganea2018hyperbolic}, (viii) Hyperbolic Graph Convolutional Neural Networks (HGCN)~\cite{chami2019hyperbolic}, (ix) Persistence Enhanced Graph Neural Network (PEGN)~\cite{zhao2020persistence}, and (x) Topological Loop-Counting Graph Neural Network (TLC-GNN)~\cite{yan2021link}. Further details of baselines are contained in Appendix B.1.2.
% convolutional GNNs (ConvGNNs)
% neural networks (NNs)
% persistent homology (PH)

{\bf Experiment Settings} We implement our proposed BScNets with Pytorch framework on two NVIDIA RTX 3090 GPUs with 24 GiB RAM. Following~\cite{chami2019hyperbolic}, for all datasets, we randomly split edges into 85\%/5\%/10\% for training, validation, and testing. Further, for all datasets, BScNets is trained by the Adam optimizer with the Cross Entropy Loss function. For all methods, we run 20 times with the same partition and report the average and standard deviation of ROC AUC values. More details about the experimental setup and hyperparameters are in Appendix B.2. Our datasets and codes are available on \url{https://github.com/BScNets/BScNets.git}.

\medskip
{\bf Experiment Results} Tables~\ref{lp_result} and~\ref{high_school_result} show the comparison of our proposed BScNets with SOAs for link prediction tasks on six network datasets (i.e., Cora, PubMed, Meetings, Phone Calls, Airport, and Disease) and two contact network datasets (i.e., High School and Staff Community), respectively. The results indicate that our BScNets consistently achieves the best performance on all network datasets compared to all SOAs. Particularly, from Table~\ref{lp_result}, we find that: (i) Compared to the spectral-based ConvGNNs (i.e., GCN and SGC), BScNets yields more than 3.32\% relative improvements to the existing best results for all six datasets; (ii) BScNets outperforms spatial-based ConvGNNs (i.e., GAT, SAGE, and SEAL) with a significant margin; (iii) Compared to the hyperbolic-based NNs (i.e., HNN and HGCN), BScNets improves upon HGCN (which performs best among the hyperbolic-based models) by a margin of 2.00\%, 1.29\%, 5.51\%, 11.62\%, 1.20\% and 7.91\% on datasets Cora, PubMed, Meetings, Phone Calls, Airport, and Disease, respectively; (iv) BScNets further improves PH-based ConvGNNs (i.e., PEGN and TLC-GNN) with a significant margin on all six datasets. Additionally, Table~\ref{high_school_result} shows  performance of BScNets and baseline methods on High School network and Staff Community. BScNets obtains the superior results on all datasets, outperforming the representative spectral-,  spatial-, PH-based models including GCN, SEAL, and TLC-GNN by a large margin. Based on the one-sided two-sample $t$-test, our BScNets also demonstrates statistically significant improvement in performance (marked by $^{\ast}$) compared to the existing best results in ROC AUC for all eight datasets (see Appendix B.5 for additional experimental results of BScNets).

Overall, the results show that BScNets can accurately capture the key
structural information on the graph, both at the dyadic and polyadic levels of interactions, and achieves a highly promising performance in link prediction.

\begin{table*}
\caption{ROC AUC and standard deviations for link prediction.  Bold numbers denote the best results. $^{\ast}$ means statistically significant result.\label{lp_result}}
\centering
\small
\setlength\tabcolsep{2pt} %5pt
\begin{tabular}{lcccccc}
\toprule
{\textbf{Model}}& {\textbf{Cora}} & {\textbf{PubMed}} & {\textbf{Meetings}} & {\textbf{Phone Calls}} & {\textbf{Airport}} & {\textbf{Disease}} 
\\
\midrule
MLP&83.15 $\pm$ 0.51 & 84.10 $\pm$ 0.97 &63.20 $\pm$ 6.22 & 60.10 $\pm$ 6.72&89.51 $\pm$ 0.52& 72.62 $\pm$ 0.61  \\
HNN~\cite{ganea2018hyperbolic}&89.00 $\pm$ 0.10 &94.87 $\pm$ 0.11 &71.00 $\pm$ 3.28& 60.90 $\pm$ 4.25& 90.78 $\pm$ 0.22& 75.10 $\pm$ 0.35\\
GCN~\cite{kipf2016semi}&90.42 $\pm$ 0.28 & 91.11 $\pm$ 0.55 &72.08 $\pm$ 4.19&61.50 $\pm$ 5.80 &89.27 $\pm$ 0.42& 64.70 $\pm$ 0.56\\
GAT~\cite{velivckovic2018graph}& 93.89 $\pm$ 0.13 & 91.22 $\pm$ 0.12 &74.00 $\pm$ 4.68& 63.40 $\pm$ 5.20 &90.55 $\pm$ 0.37&69.99 $\pm$ 0.32\\
SAGE~\cite{hamilton2017inductive}&86.24 $\pm$ 0.65 &85.96 $\pm$ 1.16&72.30 $\pm$ 5.25& 62.07 $\pm$ 5.49&90.47 $\pm$ 0.59&65.91 $\pm$ 0.33\\
SGC~\cite{wu2019simplifying}& 91.67 $\pm$ 0.20 &94.10 $\pm$ 0.20 &73.38 $\pm$ 3.49& 63.80 $\pm$ 5.71&90.01 $\pm$ 0.32& 65.21 $\pm$ 0.23 \\
SEAL~\cite{zhang2018link}&92.55 $\pm$ 0.50 & 92.42 $\pm$ 0.12 &71.09 $\pm$ 7.50& 62.96 $\pm$ 4.17& 95.16 $\pm$ 0.39 &85.23 $\pm$ 0.79\\
HGCN~\cite{chami2019hyperbolic}& 93.00 $\pm$ 0.45 & 96.29 $\pm$ 0.18 &83.20 $\pm$ 4.15 & 70.20 $\pm$ 3.77 & 96.40 $\pm$ 0.19 &90.80 $\pm$ 0.30\\
PEGN~\cite{zhao2020persistence}& 93.13 $\pm$ 0.50 & 95.82 $\pm$ 0.20 &74.17 $\pm$ 5.00 &65.23 $\pm$ 4.15&95.46 $\pm$ 0.71 &83.61 $\pm$ 1.26\\
TLC-GNN~\cite{yan2021link}& 94.22 $\pm$ 0.78 &97.03 $\pm$ 0.10 &73.20 $\pm$ 5.32 & 66.17 $\pm$ 3.90 &96.60 $\pm$ 0.69&86.19 $\pm$ 1.23\\
\midrule
\textbf{BScNets (ours)}& $^{\ast}$$\textbf{94.90}$ $\pm$ $\textbf{0.70}$ & $^{\ast}$$\textbf{97.55}$ $\pm$ $\textbf{0.12}$ & $^{\ast}$$\textbf{88.05}$ $\pm$ $\textbf{5.51}$ & $^{\ast}$$\textbf{79.43}$ $\pm$ $\textbf{6.04}$ & $^{\ast}$$\textbf{97.57}$ $\pm$ $\textbf{0.67}$ & $^{\ast}$$\textbf{98.60}$ $\pm$ $\textbf{0.58}$\\ 
\bottomrule
\end{tabular}
\end{table*}

\begin{table}[H]
\centering
%\scriptsize
\small
\caption{ROC AUC for link prediction on contact networks. \label{high_school_result}}
\resizebox{1.0\columnwidth}{!}{%%%
\begin{tabular}{lcc}
\toprule
\textbf{Model} & \textbf{High School} & \textbf{Staff Community} \\
\midrule
GCN~\cite{kipf2016semi}& 63.55 $\pm$ 1.72&64.97 $\pm$ 6.88\\
SEAL~\cite{zhang2018link}& 68.13 $\pm$ 1.50 & 65.60 $\pm$ 3.46\\
HGCN~\cite{chami2019hyperbolic}& 67.30 $\pm$ 1.28 &66.75 $\pm$ 1.38\\
TLC-GNN~\cite{yan2021link} & 69.15 $\pm$ 1.49 & 67.35 $\pm$ 5.29\\
\midrule
\textbf{BScNets (ours)}& $^{\ast}$$\textbf{71.68}$ $\pm$ $\textbf{1.72}$& $^{\ast}$$\textbf{79.13}$ $\pm$ $\textbf{2.95}$\\
\bottomrule
\end{tabular}
}
\end{table}

{\bf Ablation Study} To further investigate the importance of the different components in BScNets, we have conducted an ablation study of our proposed model on Cora and Disease and results are presented in Table~\ref{ablation_architecture} ($^{\ast}$ means statistically significant result). We compare our BScNets with three ablated variants, i.e., (i) BScNets without random walk on block Hodge-Laplacian (W/o Random walks), (ii) BScNets without relation modeling (off-diagonal terms) (W/o Relation), and (iii) BScNets without the AHLB operator but with Hodge 1-Laplacian (With only $\boldsymbol{L}_1$; i.e., instead of using the block structure, we directly incorporate Hodge 1-Laplacian $\boldsymbol{L}_1$ into the Hodge-style convolution module by replacing $\tilde{\boldsymbol{\mathbbm{L}}}^B_2$ with $\boldsymbol{L}_1$ in Eq.~\ref{h_abc_eq}). The results indicate that, when ablating the above components, the ROC AUC score of BScNets drops significantly. For both datasets, random walk mechanism on the block Hodge-style representation significantly improves the results as it utilizes higher-order relationships expressed in the graph data and learns embeddings beyond the node-space. In addition, we show that BScNets outperforms BScNets without relation modeling due to the fact that relation modeling integrates the learnt relationship between information in multi-dimensional simplices into the Hodge-style adaptive block convolutional encoder. Moreover, as expected, we find that replacing the AHLB operator by only Hodge 1-Laplacian results in a significant decrease in performance that indicates that block structure and off-diagonal higher-order relationships consistently boost the performance of link prediction.

\begin{table}[h!]
\centering
\caption{Ablation study of the network architecture (\%). \label{ablation_architecture}}
\begin{tabular}{lcc}
\toprule
\textbf{Architecture} & \textbf{Cora}& \textbf{Disease}\\
\midrule
\textbf{BScNets} & $^{\ast}$$\textbf{94.90}$ $\pm$ $\textbf{0.70}$ & $^{\ast}$$\textbf{98.60}$ $\pm$ $\textbf{0.58}$\\
W/o Random walk &93.79 $\pm$ 0.85& 93.38 $\pm$ 5.85\\
W/o Relation  &94.00 $\pm$ 0.60 & 95.90 $\pm$ 1.13\\
With $\boldsymbol{L}_1$ &93.73 $\pm$ 0.44& 93.43 $\pm$ 1.89\\
\bottomrule
\end{tabular}
\end{table}

\medskip
{\bf Computational Complexity}
Incidence matrices $\boldsymbol{B}_1$ and $\boldsymbol{B}_2$ can be calculated efficiently with the computational complexity $\mathcal{O}(n + m)$ and $\mathcal{O}(m + q)$ respectively, where $n$ is the number of 0-simplices (i.e., nodes), $m$ is the number of 1-simplices (i.e., edges), and $q$ is the number of 2-simplices (filled triangles). For large-scale graphs, we sample $\epsilon \times m$ edges (where $\epsilon \in (0 , 1]$) from edge set $\mathcal{E}$ and then pass these $\epsilon \times m$ edges to construct higher-order Hodge Laplacians. Thus this sparse sampling reduces the computation complexity of constructing $\boldsymbol{B}_1$ and $\boldsymbol{B}_2$ to $\mathcal{O}(n + \epsilon \times m)$ and $\mathcal{O}(\epsilon \times m + q^\prime)$ respectively (where $q^\prime$ denotes the number of 2-simplices after sparse sampling). This sampling method is motivated by the DropEdge technique~\cite{rong2019dropedge} which has been used to prevent over-fitting and over-smoothing in GCNs via randomly removing edges from the graph. Equipping BScNets with edge sampling results in substantial reductions
in the size of the input data and, as such, computational complexity. %Remarkably, while in the case of large networks BScNets operates only on the 10\% of the input data other methods are allowed to use, BScNets still significantly outperforms SOA. 
Table in Appendix B.4 reports the average running time of incidence matrices generation for target network, training time per epoch of our BScNets model on all datasets, and running time comparison between BScNets and baselines.

\medskip
{\bf Infectious Disease Transmission}
As recently noted by~\citet{iacopini2019simplicial, schaub2020random, li2021contagion}, higher-order relationships such as multi-member group interactions, may be the key driving factors behind contagion dynamics,
and simplicial complexes offer a natural way to describe such polyadic group relations. Here we are motivated by the two interlinked questions: {\it Can we use link prediction with and without simplicial complexes to reconstruct unobserved social interactions and then to approximate the underlying contagion dynamics in the whole population? How accurate are the link prediction methods, with and without simplicial complexes, in reflecting the impact of disease mitigation strategies?}

Indeed, in practice public healthcare professionals might have records on most registered residents (i.e., nodes) via, e.g., social security offices. However,   
there is only partial information on social interactions (i.e., edges) among these individuals. Our goals are to (i) predict social links among individuals where the interaction information is unknown, (ii) assess how mitigation strategies will impact the epidemic spread predicted from the reconstructed network, and (iii) evaluate how close the infection curves on the reconstructed data are to the infection curve on the whole true population. More details about the procedure of infectious disease transmission evaluation are provided in Appendix B.3.

%Our goal is then to reconstruct social links among individuals and to evaluate how close the infection curve on the reconstructed data is to the infection curve on the whole true population. 

We adopt the Susceptible-Exposed-Infected-Recovered (SEIR)~\cite{kermack1927contribution} epidemic model. We set the infection curve delivered by SEIR on the whole population as the ground truth. We perturb certain fraction of links (i.e., remove the real links and add fake links) from the whole population to address the real world scenario when we do not have access to information on all social interactions. We then apply two link prediction methods to the network with perturbation, with and without simplicial complexes, BScNets and TLC-GNN, respectively. We now reconstruct the whole population by using BScNets and TLC-GNN and consider the two infection curves yielded by SEIR on the BScNets-based and TLC-GNN-based reconstructed networks. In addition, we assess sensitivity of the BScNets and TLC-GNN curves to mitigation strategies where more central nodes (individuals) are targeted to receive a vaccine, to quarantine or are persuaded to wear masks~\cite{chen2021prioritizing, curiel2021vaccination}. 
Under the targeted mitigation strategies, a fraction of the most central nodes cannot transmit the disease. The `base' infection curve is obtained from the whole population where such central nodes are removed from the disease spread. In turn, BScNets and TLC-GNN operate on the partially observed data where some edges are perturbed (as discussed above) and, upon reconstructing the unknown social interactions, BScNets and TLC-GNN are asked to re-determine their own individuals to be targeted through the disease mitigation strategy. We perform these experiments on the High School network~\cite{salathe2010high, eletreby2020effects}. We select TLC-GNN as the competing link prediction approach as it is a runner-up  (see Table~\ref{high_school_result}). We set the edge perturbation rate to 20\%. We consider betweenness and degree centralities. We simulate each scenario 50 times and average the results.
(For more experiments see Appendix B.3).

%Since in practice we never observe all social interactions among humans and have access only to a small portion of the data, we perform subsampling of the population (with a rate of \%??).
%We then apply two link prediction methods to these subsampled data, with and without simplicial complexes, BScNets and TLC-GNN, respectively. We now reconstruct the whole population using BScNets and TLC-GNN and consider the two infection curves yielded by SEIR on the BScNets-based and TLC-GNN-based reconstructed networks. 

%To investigate the feasibility of our proposed BScNets for real-world problems, we conduct additional analyses for understanding the dynamics of infectious disease spread (e.g., COVID-19) on the social network, i.e., High School network. The classical Susceptible-[Exposed]-Infected-Recovered (SIR/SEIR)~\cite{kermack1927contribution} epidemic models have been widely used in the study of simulating the epidemic dynamics. In this paper, we consider adopting SEIR for modeling the process of the epidemic based on predicted network via link prediction of each model. The SEIR model has four compartments which are $S$ (susceptible), $E$ (exposed), $I$ (infectious), and $R$ (recovered). In our study, we assume the SEIR model with spreading process that starts with one infected individual and 20\% of individuals adopted mitigation strategy (e.g., mask-wearing and vaccination doses administered). More detailed parameters for SEIR are deferred to Appendix B.2. 

\begin{figure}[H] % [h!]
    \centering
    \includegraphics[width=0.35\textwidth]{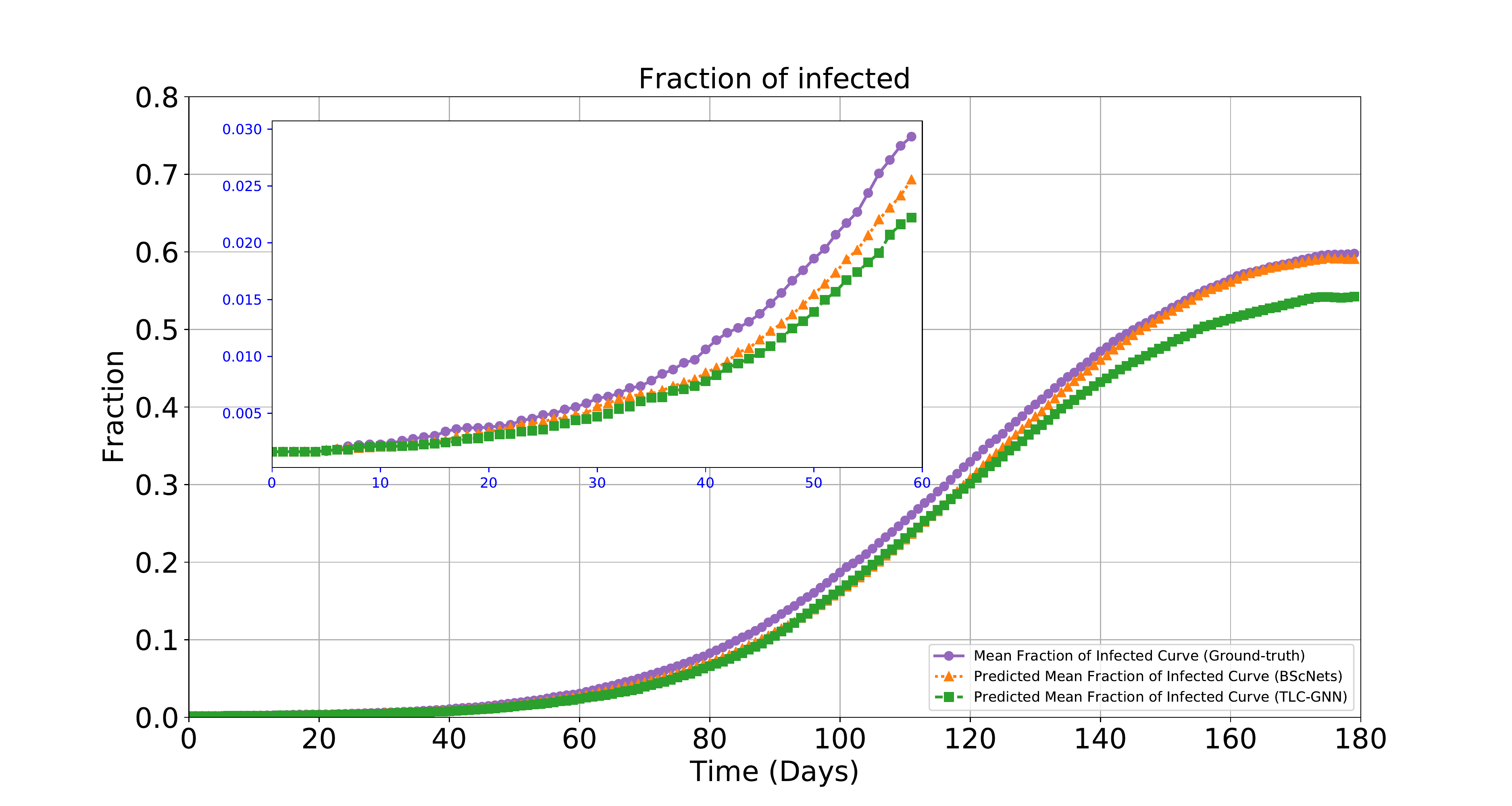}
    \caption{SEIR infection curves
    based on the original High School (HS) network ({\color{black}{\it purple}}), BScNets-reconstructed HS network({\color{black}{\it orange}}), and TLC-GNN-reconstructed HS network ({\color{black}{\it green}}), under betweenness-based mitigation strategy.\label{epi_curves}}
\end{figure}

Figure~\ref{epi_curves} shows the infection curves during the time period of 180 days fitted to (i) the original network, (ii) the BScNets reconstructed network, and (iii) the TLC-GNN reconstructed network, under targeted mitigation strategy based on the betweenness centrality. We observe that the infection curve of the BScNets-reconstructed network (curve\textsubscript{BScNets}) is significantly closer to the `base' infection curve (i.e., curve\textsubscript{base}) than the curve based on TLC-GNN (curve\textsubscript{runner-up}). Most importantly, we find that 
while the peaks of the curve\textsubscript{BScNets} and the curve\textsubscript{base} are very close, i.e. 59.1\% and 59.8\%,
respectively,
%365 and 369 infected individuals, ,
the TLC-GNN curve (which does not account for multi-node group interactions) tends to 
substantially underestimate the number of infected individuals. For instance,
at day $t = 30$, $t = 60$, $t = 90$, $t = 120$, and at day $t = 180$, 
TLC-GNN suggests that only 0.5\%, 2.2\%, 10.1\%, 29.5\%, and 54.2\% of population are infected, while the ground truth `base' curve projects that 0.6\%, 2.9\%, 12.2\%, 32.2\%, and 59.8\% are infected. That is, the difference between TLC-GNN and the `base' curve, even for such small network, reaches 5\% of the population, which eventually translates into significant societal implications such as shortages of health care facilities and overall system unpreparedness to pandemics like COVID-19.
This phenomenon also confirms the most recent premise of epidemiological studies that higher-order interactions among multi-node groups (i.e., simplicial complexes) are the hidden driving factors behind the disease transmission mechanism. In turn, BScNets and other SNNs may be the most promising direction not only to capture such higher-order group interactions but also to reveal hidden polyadic relations in social networks.

%4, 18, 76, 199, and 369 infected individuals respectively while the ground truth curve expects to have 3, 14, 62, 182, and 335 infected individuals respectively.

%($\text{Peak}\textsubscript{BScNets} = 59.08\%$) is similar to the curve\textsubscript{base} ($\text{Peak}\textsubscript{base} = 59.78\%$) whereas the peak of the curve\textsubscript{runner-up} is {\it under-estimated}, i.e., much lower than ground-truth ($\text{Peak}\textsubscript{runner-up} = 54.23\%$).  

%(iii) TLC-GNN tends to delay the estimated infected individuals (around 1-day) when compared to the curve\textsubscript{base} which is the potential loss for governments to take the most optimal policy actions at public agencies.

\section{Conclusion}
We have proposed the first SNN for link prediction. We have shown that relations among multiple multi-node structures play a significant role in graph learning. In the future, we will extend the ideas of simplicial DL to dynamic networks.

\section{Acknowledgements}

This material is based upon work sponsored by 
the National Science Foundation under award numbers ECCS 2039701, ECCS 2039716, INTERN supplement for ECCS 1824716, DMS 1925346 and the Department of the Navy, Office of Naval Research under ONR award number N000142112226. Any opinions, findings, and
conclusions or recommendations expressed in this material are
those of the author(s) and do not necessarily reflect the views
of the Office of Naval Research. 
%Any opinion, findings, and conclusions or recommendations expressed in this material are those of the author(s) and do not necessarily reflect the views of the National Science Foundation.

% Use \bibliography{yourbibfile} instead or the References section will not appear in your paper
\bibliography{blockscn}

\clearpage
% appendix
\section{A. Methodology}
\subsection{A.1. Notations and Simplicial Complex Visualization}
We summarize the main notations and their definitions in Table~\ref{notations}.
\begin{table}[ht!]
\centering
\caption{The main symbols and definitions in this paper.\label{notations}}
\begin{tabular}{lc}
\toprule
\textbf{Notation} &  \textbf{Definition} \\
\midrule
$\boldsymbol{A}$ & adjacency matrix\\
$\boldsymbol{X}$ & node feature matrix\\
$\boldsymbol{L}_0$ & unnormalized graph Laplacian \\
$\boldsymbol{L}^{up}_1$ & 1-up-Laplacian\\
$\boldsymbol{L}^{down}_1$ & edge Laplacian/1-down-Laplacian\\
$C^k$ & vector space with the oriented $k$-simplices\\
$\partial_k$ & boundary operator\\
$\boldsymbol{B}_k$ & incidence matrix/matrix representation of $\partial_k$\\
$\boldsymbol{L}_k$ & $k$-Hodge Laplacian\\
$\boldsymbol{\mathfrak{L}}_K$ & $K$-block Hodge-Laplacian\\
$^R\boldsymbol{\mathfrak{L}}_J$ & reduced block Hodge-Laplacian\\
$\boldsymbol{\mathfrak{L}}_2$ & 2-block (reduced) Hodge-Laplacian\\
$\mathbb{P}_{d_2}$ & orthogonal projector\\ 
$\boldsymbol{\mathbbm{L}}^{B}_2$ & adaptive hodge Laplacian based block (AHLB)\\
\bottomrule
\end{tabular}
\end{table}

\begin{figure}[ht!]
    \centering
    \includegraphics[width=0.4\textwidth]{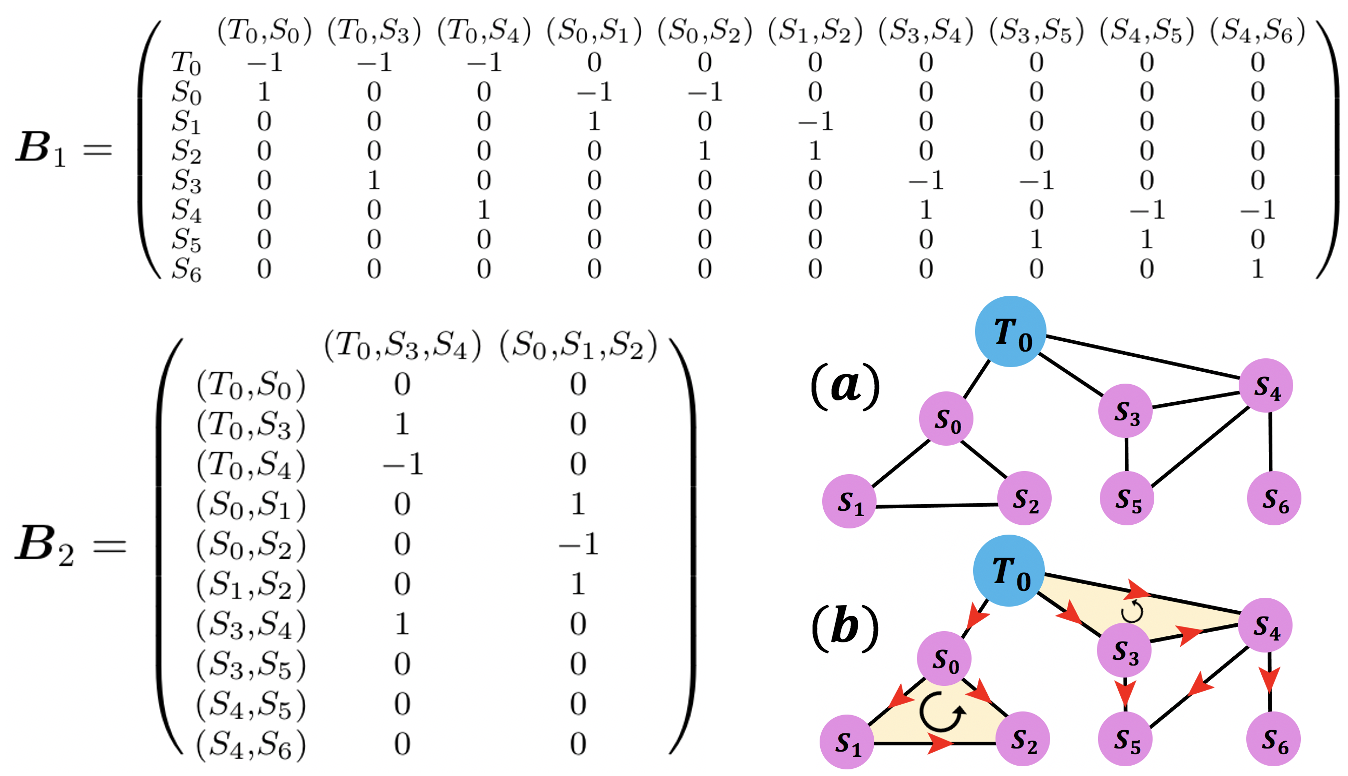}
    \caption{An example of generating simplicial complex from a classroom environment network, where $T_0$ is the teacher\_0 and $S_u$ is the student\_$u$ ($u \in \{0, \dots, 6\}$). (a) Graph structure of the classroom environment network; (b) Simplicial complex of the classroom environment network; $\boldsymbol{B}_1$ and $\boldsymbol{B}_2$ are the corresponding node-to-edge and edge-to-face incidence matrices, respectively.}
    \label{B1_B2_illustration_appendix}
\end{figure}

We visualize an example of synthetic classroom environment network and its corresponding simplicial complex in Figure~\ref{B1_B2_illustration_appendix}. The node $T_0$ (blue node) represents the teacher\_0 and node $S_u$ (pink node) represents a student\_$u$. The shaded areas represent the 2-simplices (triangles), i.e., $\{S_0, S_1, S_2\}$ and $\{T_0, S_3, S_4\}$. Besides, we show the orientation of each simplex graphically using arrows. 

\subsection{A.2. Lemmas and Proofs}

\begin{lemma}
For $K, J\in \mathbb{Z}_{\geq 0}$, operators $\boldsymbol{\mathfrak{L}}_K$ and $^R\boldsymbol{\mathfrak{L}}_J$ are symmetric and semipositive-definite, i.e. $\boldsymbol{\mathfrak{L}}_K\geq 0$ and  $^R\boldsymbol{\mathfrak{L}}_J \geq 0$;
 $\boldsymbol{\mathfrak{L}}_K=(\boldsymbol{\mathfrak{L}}_{K})^{\top}$
and $^R\boldsymbol{\mathfrak{L}}_J=(^R\boldsymbol{\mathfrak{L}}_J)^{\top}$.
\end{lemma}

{\bf Proof} Since for any nonnegative integer $i$, $\boldsymbol{L}_i=\boldsymbol{L}_i^{\top}$, by construction $\boldsymbol{\mathfrak{L}}_K=\boldsymbol{\mathfrak{L}}_K^{\top}$ and $^R\boldsymbol{\mathfrak{L}}_J=^R\boldsymbol{\mathfrak{L}}_J^{\top}$.

Positive semidefiniteness of $\boldsymbol{\mathfrak{L}}_K$ and $^R\boldsymbol{\mathfrak{L}}_J$ follows from the theorem on spectra of block upper/lower triangular partitioned matrices (see~\citet{simovici2012linear}).

\begin{lemma}\label{lemma_power_appendix}
For any $r\in \mathbb{Z}_{>0}$ and $K, J\in \mathbb{Z}_{\geq 0}$, the $r$-power of 
$(\boldsymbol{\mathfrak{L}}_K)^r$ and $(^R\boldsymbol{\mathfrak{L}}_J)^r$
are given by 
$\oplus_{k=0}^K ((\boldsymbol{L}^{down}_k)^r+(\boldsymbol{L}^{up}_k)^r)$
and $\oplus_{j=1}^J ((\boldsymbol{L}^{down}_{k_j})^r+(\boldsymbol{L}^{up}_{k_j})^r)$, respectively.
\end{lemma}

{\bf Proof} In view of lemma~33 of~\citet{bodnar2021weisfeiler}, for each nonnegative integers $r$ and $k$,  
$\boldsymbol{L}_k^r=(\boldsymbol{L}^{down}_k+\boldsymbol{L}^{up}_k)^r=(\boldsymbol{L}^{down}_k)^r+(\boldsymbol{L}^{up}_k)^r$.  
The result of Lemma~2 then follows from the multiplication rule for block matrices~\cite{eves1980elementary}.

\medskip
Consider the adaptive Hodge Laplacian Based (AHLB) Block operator 
\begin{align}
\label{ahlb_laplacian_appendix}
\boldsymbol{\mathbbm{L}}^{B}_2=\left[
\begin{array}{c|c}
\boldsymbol{L}_{k_1} & f(\mathbb{P}_{d_{2}}\boldsymbol{L}_{k_1}, \boldsymbol{L}_{k_2}) \\ \hline
f(\mathbb{P}_{d_{2}}\boldsymbol{L}_{k_1}, \boldsymbol{L}_{k_2})^{\top} & \boldsymbol{L}_{k_2}
\end{array}\right],
\end{align}
where the off-diagonal terms are defined via the embedded Inner-Product
$$f([\mathbb{P}_{d_{2}}\boldsymbol{L}_{k_1}]_i, [\boldsymbol{L}_{k_2}]_j) =
<\boldsymbol{\Theta}_{\xi} [\mathbb{P}_{d_{2}}\boldsymbol{L}_{k_1}]_i, \boldsymbol{\Theta}_{\psi}[\boldsymbol{L}_{k_2}]_j>.
%(\boldsymbol{\Theta}_{\xi} [\mathbb{P}_{d_{2}}\boldsymbol{L}_{k_1}]_i)^{\top} \cdot \left(\boldsymbol{\Theta}_{\psi}[\boldsymbol{L}_{k_2}]_j\right).
$$
Here $\boldsymbol{\Theta}_{\xi} \in \mathbb{R}^{d_c \times d_2}$ and $\boldsymbol{\Theta}_{\psi} \in \mathbb{R}^{d_c \times d_2}$ are weight matrices to be learned, and $d_c$ is the embedding dimension. 
Denote a $d_c\times d_2$-matrix $\boldsymbol{\Theta}_{\xi} [\mathbb{P}_{d_{2}}\boldsymbol{L}_{k_1}]$ as $\boldsymbol{W}$
and a $d_c\times d_2$-matrix $\boldsymbol{\Theta}_{\psi} \boldsymbol{L}_{k_2}$ as $\boldsymbol{U}$. Hence, 
since we consider Laplacians over the real field $\mathbb{R}$ and, as such, can use dot product, 
$\boldsymbol{\mathbbm{L}}^{B}_2$ can be re-written as
\begin{align}
\label{ahlb_laplacian_rev}
\boldsymbol{\mathbbm{L}}^{B}_2=\left[
\begin{array}{c|c}
\boldsymbol{L}_{k_1} & \boldsymbol{W}^{\top}\boldsymbol{U}\\ \hline
\boldsymbol{U}^{\top}\boldsymbol{W} & \boldsymbol{L}_{k_2}
\end{array}\right].
\end{align}

\begin{lemma} Given the adaptive Hodge Laplacian Based (AHLB) Block operator $\boldsymbol{\mathbbm{L}}^{B}_2$,
the following conditions are equivalent:
\begin{enumerate}
\item $\boldsymbol{\mathbbm{L}}^{B}_2\geq 0$,
\item  $\boldsymbol{U}^{\top}\boldsymbol{W}=\boldsymbol{L}_{k_2}\boldsymbol{L}_{k_2}^{\dagger}\boldsymbol{U}^{\top}\boldsymbol{W}$ and \\
$\boldsymbol{L}_{k_1}-\boldsymbol{W}^{\top}\boldsymbol{U} \boldsymbol{L}_{k_2}^{\dagger}\boldsymbol{U}^{\top}\boldsymbol{W}\geq0$,

\item $\boldsymbol{W}^{\top}\boldsymbol{U}- \boldsymbol{L}_{k_1}\boldsymbol{L}_{k_1}^{\dagger} \boldsymbol{W}^{\top}\boldsymbol{U}$ and \\
$\boldsymbol{L}_{k_2}-\boldsymbol{U}^{\top}\boldsymbol{W}\boldsymbol{L}_{k_1}^{\dagger}\boldsymbol{W}^{\top}\boldsymbol{U}\geq 0$,
\end{enumerate}
where 
$\boldsymbol{U}=\boldsymbol{\Theta}_{\psi} \boldsymbol{L}_{k_2}$,
$\boldsymbol{W}=\boldsymbol{\Theta}_{\xi} [\mathbb{P}_{d_{2}}\boldsymbol{L}_{k_1}]$, and $\dagger$ denotes the Moore-Penrose inverse of a matrix. Here
$\boldsymbol{L}_{k_1}-\boldsymbol{W}^{\top}\boldsymbol{U} \boldsymbol{L}_{k_2}^{\dagger}\boldsymbol{U}^{\top}\boldsymbol{W}$ and $\boldsymbol{L}_{k_2}-\boldsymbol{U}^{\top}\boldsymbol{W}\boldsymbol{L}_{k_1}^{\dagger}\boldsymbol{W}^{\top}\boldsymbol{U}$ are called Schur complements.
\end{lemma}

{\bf Proof} Since $\boldsymbol{L}_{k_2}\geq 0$, $\boldsymbol{L}_{k_1}=\boldsymbol{L}_{k_1}^{\top}$ and $\boldsymbol{L}_{k_2}=\boldsymbol{L}_{k_2}^{\top}$, condition~3) is equivalent to condition i) of Theorem~1 by~\citet{albert1969conditions} (see p.435). Hence, by the Albert theorem, condition 3) holds iff $\boldsymbol{\mathbbm{L}}^{B}_2\geq 0$, that is, conditions 1) and 3) are equivalent.

In turn, since also $\boldsymbol{L}_{k_1}\geq 0$, from the Schur complement theorem, conditions 2) and 3) are equivalent (see p.263 of~\citet{bekker1988positive}), which implies that conditions 1) and 2) are equivalent and concludes the proof.

%\medskip
%Note that in general, $\boldsymbol{\mathbbm{L}}^{B}_2$ is neither necessarily positive semidefinite nor diagonally
%dominant. As such, $\boldsymbol{\mathbbm{L}}^{B}_2$ is not formally a Laplacian. Nevertheless, if needed, it is possible to impose conditions on  $\boldsymbol{\Theta}_{\psi}$ and $\boldsymbol{\Theta}_{\xi}$ to ensure that $\boldsymbol{\mathbbm{L}}^{B}_2$ satisfies these requirements.
%\YGL{Yuzhou, do we need the paragraph above?}  
%\YC{Yes, I think we can keep it.}

\subsection{A.3. Random walk-based $K$-block Hodge-Laplacian}
In our experiments, followed by Lemma 2 (see main body part), we consider random walks on multiple simplices by generalizing Eq. 3 (see main body part) to arbitrary power $r$ (where $r \in \mathbb{Z}_{>0}$) gives
\begin{equation*}
\boldsymbol{\mathbbm{L}}^r_2=\left[
\begin{array}{c|c}
\boldsymbol{L}^r_{k_1} & f(\mathbb{P}_{d_{2}}\boldsymbol{L}^r_{k_1}, \boldsymbol{L}^r_{k_2})\\
\hline f(\mathbb{P}_{d_{2}}\boldsymbol{L}^r_{k_1}, \boldsymbol{L}^r_{k_2})^{\top} & \boldsymbol{L}^r_{k_2}
\end{array}\right].
\end{equation*}
\subsection{A.4. The overall framework of BScNets}
\begin{figure}[h!] % [t!]
    \centering
    \includegraphics[width=0.47\textwidth]{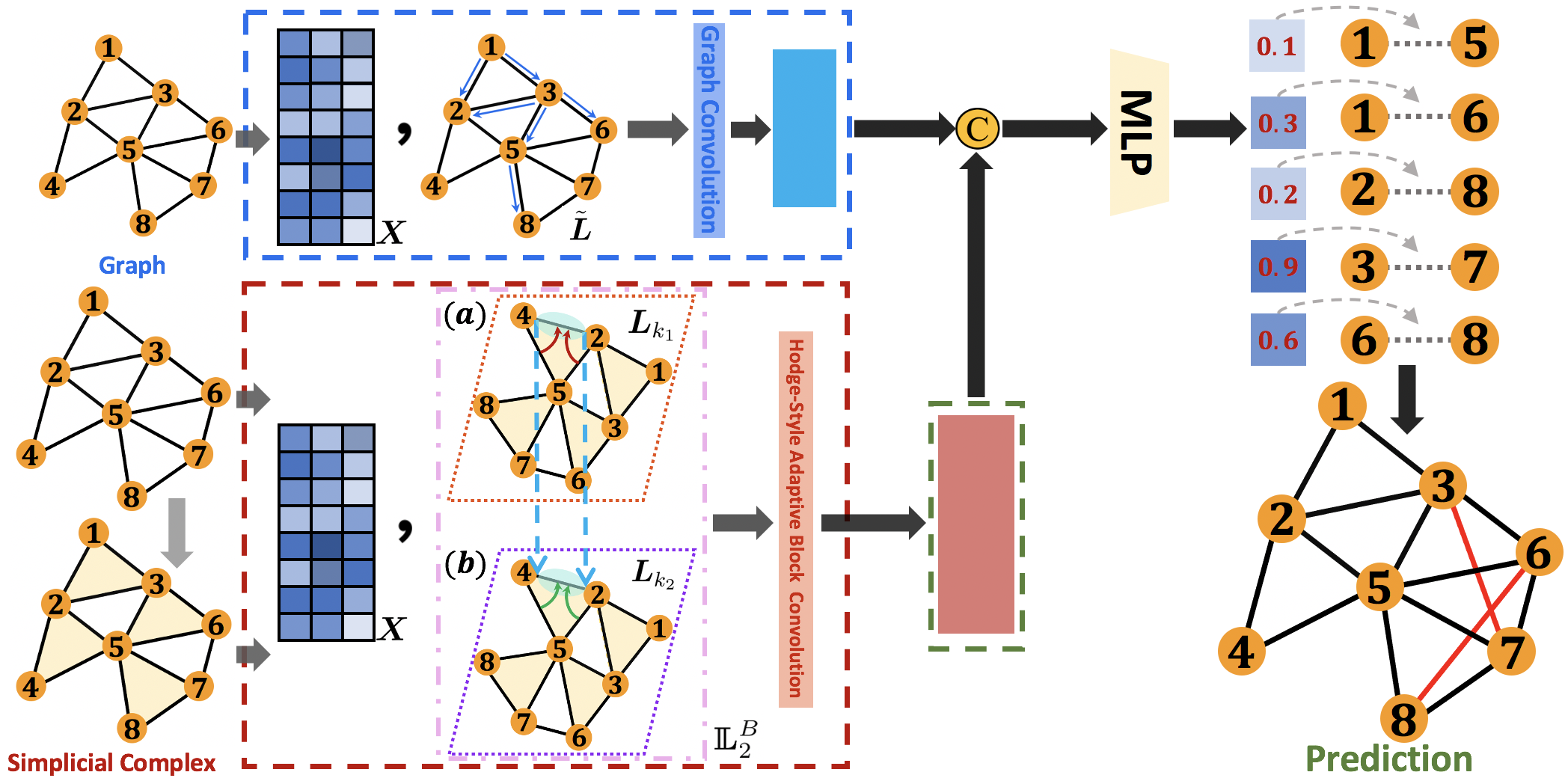}
    \caption{The architecture of BScNets. BScNets consists of graph convolution (upper row) and Hodge-style adaptive block convolution module (lower row). Each convolution layer/module contains the node feature matrix $\boldsymbol{X}$ and a Laplacian/operator (i.e., normalized Laplacian $\tilde{\boldsymbol{L}}$ and adaptive Hodge Laplacian based operator $\boldsymbol{\mathbbm{L}}^{B}_2$ for upper and lower parts respectively). Lower row presents the relation modeling between (a) Hodge Laplacian $\boldsymbol{L}_{k_1}$ describing diffusion from edge to edge through nodes and (b) Hodge Laplacian $\boldsymbol{L}_{k_2}$ describing diffusion from edge to edge through triangles. Symbol $\circled{C}$ represents concatenation.\label{flowchart_appendix}}
\end{figure}

The details of the architecture of our proposed BScNets are as follows. Figure~\ref{flowchart_appendix} shows that in BScNets  (i) the graph convolution is employed directly on graph-structured data to model node embeddings by aggregating the embeddings of neighbors and (ii) the Hodge-style adaptive block convolution module is constructed to jointly model edge embeddings via (1) message passing from edges to edges through nodes, (2) massage passing from edges to edges through triangles, and (3) relation modeling among Hodge representations from two simplicies. Lastly, we concatenate two corresponding embeddings and deploy a MLP to predict whether an edge exists between any pair of nodes.

\section{B. Detailed Experimental Settings}
\subsection{B.1. Datasets and Baselines}
\subsubsection{B.1.1 Datasets}
We experiment on three types of networks (i) citation networks: Cora and PubMed, (ii) social networks: Airport, Meetings, Phone Calls, High School network, Teacher Community, Staff Community, and (iii) disease propagation tree: Disease. Statistical overview of all eight datasets is in Table~\ref{dataset}. We now provide the detailed descriptions of these three types of datasets we used as follows (i) Citation networks (i.e., Cora and Pubmed~\cite{sen2008collective}) represent the citation relationship between two papers (where nodes represent papers and edges denote citation links) and node features are Bag-of-Words (BoW) representations of the papers. (ii) For social networks, (1) Airport~\cite{chami2019hyperbolic} is a flight network dataset from \url{openflights.org} where nodes represent airports and edges represent the airline routes, (2) Meetings and Phone Calls are real criminal datastes built from juridical acts~\cite{cavallaro2020disrupting}: the Meetings network represents the physical meetings among members and the Phone Calls network represents phone calls between individuals (where nodes represent, e.g., members in the ``Mistretta'' family and the ``Batanesi'' clan and edges represent meetings or phone calls exchanged between pairs of individuals), (3) High School network and Staff Community: High School is the close proximity interaction (CPI) network observed at a U.S. high school during a typical day~\cite{salathe2010high} where nodes represent students, teachers, and staffs and edges represent interactions between two individuals; Staff Community is subnetwork contains nodes with staff label; following~\cite{eletreby2020effects}, we note that we sample a static graph out of the raw High School network by assigning an edge with the ratio of the total contact time between two nodes to the maximum total contact time observed in the dataset; it is noteworthy that social networks (i.e., Meetings, Phone Calls, High School network, and Staff Community) have no node attributes and thus we select four classic centralities as node features including degree, closeness centrality, betweenness centrality, and pagerank centrality. (iii) Disease propagation tree (Disease)~\cite{chami2019hyperbolic} are simulated from the SIR model~\cite{anderson1992infectious}, where node attributes indicate the susceptibility to the disease and are generated through random sampling of Gaussian distributions (i.e., the mean and standard deviation are set for susceptible to disease and non-susceptible to disease, respectively). Table~\ref{dataset} summarizes the statistics of the datasets.

\begin{table}[ht!]
\centering
\caption{Summary of datasets used in link prediction tasks.\label{dataset}}
\begin{tabular}{lccc}
\toprule
\textbf{Dataset} &  \textbf{\# Nodes} & \textbf{\# Edges} & \textbf{\# Features} \\
\midrule
Cora & 2,708& 5,429 &1,433\\
PubMed &19,717 &44,338 &500\\
Meetings &101 &256 &4\\
Phone Calls &100 &124 &4\\
Airport &3,188 &18,631 &4\\
Disease &1,044 &1,043 &1,000\\
High School & 773 & 6,342 & 4\\
%Teacher Community &73 &519 &4 \\ % or federation of teachers
Staff Community &55 &441 &4 \\
\bottomrule
\end{tabular}
\end{table}

\subsubsection{B.1.2 Baselines}
We compare against ten state-of-the-art (SOA) baselines, including (i) MLP: which is a feature-based approach to make link prediction without utilizing the graph structure information; (ii) two spectral-based convolutional GNNs (ConvGNNs): (1) GCN~\cite{kipf2016semi} and (2) Simplified Graph Convolution (SGC)~\cite{wu2019simplifying} which simplifies the multi-layer graph convolutional networks by multiplying the adjacency matrix for $k$ times; (iii) three spatial-based ConvGNNs: (1) Graph Attention Networks (GAT)~\cite{velivckovic2018graph} which adopts attention mechanism to aggregate the information from neighbors' representation investigating the relative weights between two connected nodes, (2) GraphSAG (SAGE) which learns the node representations through sampling and aggregating features from its local neighborhood, and (3) SEAL~\cite{zhang2018link} which predicts the relationship on graph-structure data based on graph convolution on local subgraph around each link; (iv) two hyperbolic-based NNs: (1) Hyperbolic neural networks (HNN)~\cite{ganea2018hyperbolic} which is the hyperbolic version of MLP and (2) Hyperbolic Graph Convolutional Neural Networks (HGCN)~\cite{chami2019hyperbolic} which generalizes the graph convolution into hyperbolic space with aggregation in the tangent space; (v) two persistent homology (PH)-based ConvGNNs: (1) Persistence Enhanced Graph Neural Network (PEGN)~\cite{zhao2020persistence} which embeds topological information into the aggregation function and (2) Topological Loop-Counting Graph Neural Network (TLC-GNN)~\cite{yan2021link} which injects the pairwise topological information into the latent representation of a GCN.
\subsection{B.2. Hyperparameter Settings and Training Details}
BScNets consists of a one adaptive block Hodge convolution layer ({\it nhid1}) and two graph convolution layers ({\it nhid2, nhid3}), where {\it nhid1} $\in \{1, 8, 16, 32, 64, 128\}$, {\it nhid2} $\in \{64, 128, 1024\}$, and {\it nhid3} $\in \{4, 16, 32\}$. In addition, we perform an extensive grid search for learning rate among $\{0.001, 0.005, 0.008, 0.01, 0.05\}$, the dropout rate among $\{0.1, 0.2, \dots, 0.9\}$, power $r$ for random walk among $\{2, 3, 4, 5\}$, and importance weights\footnote{importance weights for distances (i.e., $\text{dist}^{\text{GC}}_{uv}$ and $\text{dist}^{\text{H-ABC}}_{uv}$; see Eqs. 6 and 8 in main body) between two nodes from graph convolution operator and Hodge-style adaptive block convolution module.} $\pi_{\alpha}$ among $\{0.01, 0.1, 1.0, 10.0, 100.0\}$ and $\pi_{\beta}$ among $\{1.0, 10.0\}$ respectively. The model is trained for 5,000 epochs with early stopping applied when the metric (i.e., validation loss) starts to drop. We use ROC AUC as the metric for deriving the best hyperparameters and it is computed via cross-validation by the grid search. Furthermore, in Appendix B.7, we provide results of hyparameters sensitivity Analysis of $\delta$ and $\eta$.
%For Cora, PubMed, Airport, High School, and Staff Community, we apply the sparse sampling approach on their edge sets with the sampling percentage $\epsilon_{\text{Cora}} = 0.2$, $\epsilon_{\text{PubMed}} = 0.01$, $\epsilon_{\text{Airport}} = 0.05$, $\epsilon_{\text{High School}} = 0.3$, and $\epsilon_{\text{Staff Community}} = 0.8$ respectively.

\subsection{B.3. Infectious Disease Transmission}
We adopt the {\bf S}usceptible-{\bf E}xposed-{\bf I}nfected-{\bf R}ecovered
(SEIR) epidemic model to investigate infectious disease transmission on High School network. In our simulation study, the parameters for SEIR model are as follows (i) transition rate ($S \rightarrow E$) $\beta = 0.01$, (ii) transition rate ($E \rightarrow I$) $\alpha = 0.1$, and (iii) recovery rate ($I \rightarrow R$) $\gamma = 0.005$. For both betweenness- and degree-based mitigation strategies, the nodes are selected in the decreasing order of their corresponding betweenness centrality or degree centrality scores. In our experiments, we set 20\% of nodes (i.e., individuals) under betweenness- or degree-based mitigation strategy, i.e., we remove 20\% of nodes with highest betweenness centrality or degree centrality scores from the targeted network.

\begin{figure}[ht!] % [h!]
    \centering
    \includegraphics[width=0.43\textwidth]{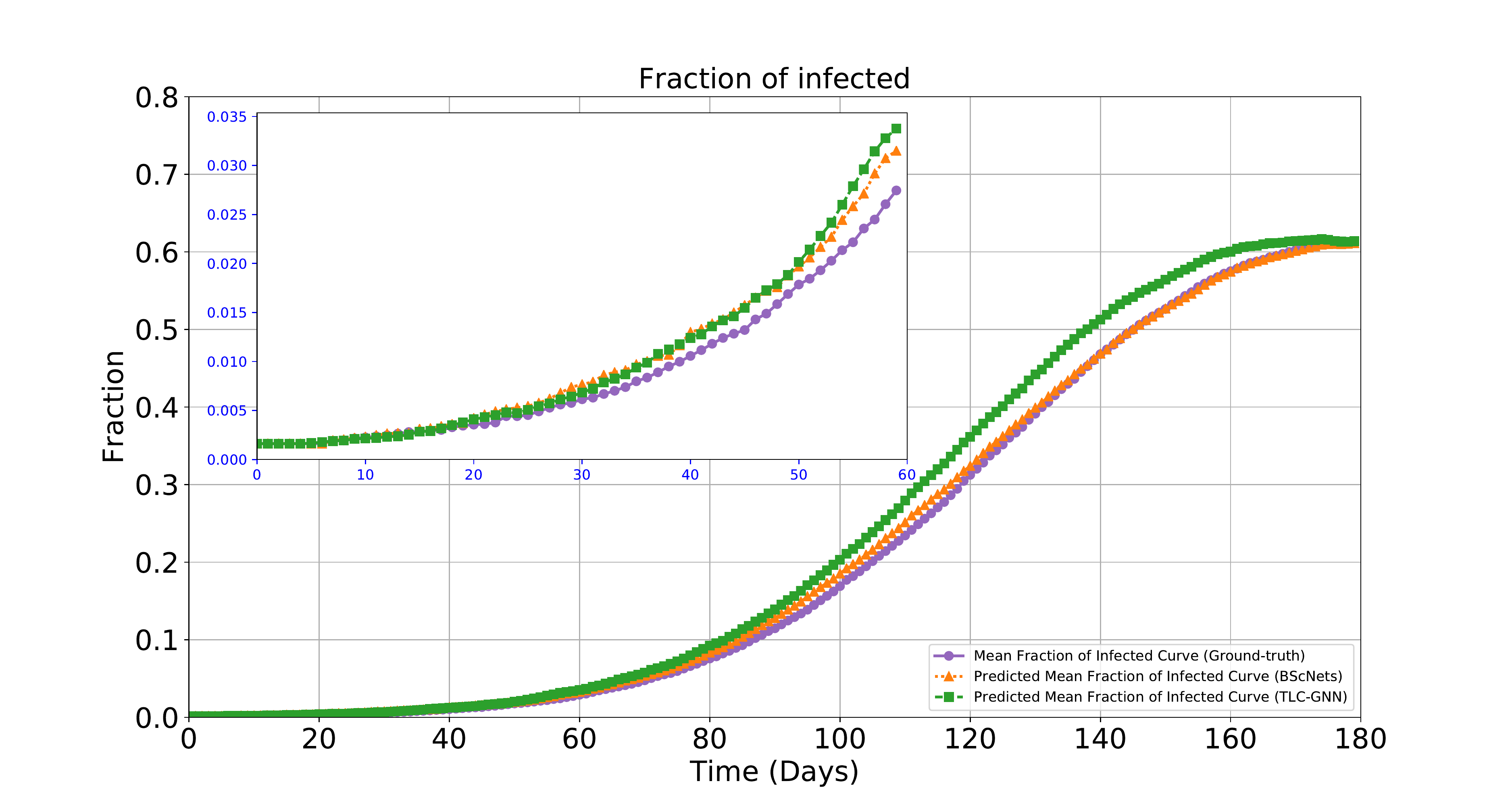}
    \caption{SEIR infection curves
    based on the original High School (HS) network ({\color{black}{\it purple}}), BScNets-reconstructed HS network({\color{black}{\it orange}}), and TLC-GNN-reconstructed HS network ({\color{black}{\it green}}), under betweenness-based mitigation strategy.\label{epi_curves_degree_based}}
\end{figure}

Besides, we also consider applying degree-based mitigation strategy on the original network, BScNets reconstructed network, and TLC-GNN reconstructed network respectively. Figure~\ref{epi_curves_degree_based} shows the infection curves during the time period of 180 days fitted to (i) the original network, (ii) the BScNets reconstructed network, and (iii) the TLC-GNN reconstructed network, under targeted mitigation strategy based
on the degree centrality. We find that the infection curves from the original network and BScNets reconstructed network are very close; however, the TLC-GNN curve (which does not
account for multi-node group interactions) tends to substantially
overestimate the number of infected individuals. For instance, at day $t = 30$, $t = 60$, $t = 90$, $t = 120$, and at day $t = 180$, 
TLC-GNN suggests that only 0.6\%, 2.7\%, 11.1\%, 30.5\%, and 61.2\% of population are infected, while the ground truth `base' curve projects that 0.6\%, 3.4\%, 13.4\%, 35.4\%, and 61.4\% are infected. This observation indicates that the decision-making (based on the estimation and forecasting from TLC-GNN) may result in time consuming and resource wasting for governments and public agencies.

\subsection{B.4. Computation Complexity}
Table~\ref{running_time} shows the computational costs for incidence matrices generation and training time per epoch of our BScNets on all eight datasets. We have shown that our BScNets equipped with the sparse sampling technique is much more efficient in terms of running time. For instance, the number of edges in PubMed is more than 16 times larger than the number of edges in Cora; however, the incidence matrices generation of PubMed exceeds the incidence matrices generation of Cora with more than 165 less computation cost (running time). In summary, our BScNets model can be conducted on large-scale real-world datasets (static and dynamic) without the issues running of out-of-memory and prohibitive computational costs. Besides that, we then further report the running time (in seconds) of BScNets and 4 baselines on Cora and Disease. We can see that BScNets outperforms baselines and maintains comparable running time.

\begin{table*}[h!]
%\captionsetup{font=Large}
\centering
\setlength\tabcolsep{3pt}
\caption{Link prediction in ROC AUC on Citeseer.\label{citeseer_res}}
%\vspace{-2ex}
\begin{tabular}{lcccccccc}
\toprule
\textbf{Dataset} & \textbf{BScNets} & \textbf{DeepWalk}&  \textbf{Node2Vec} & \textbf{Node2Vec-GCN}&  \textbf{GCN} & \textbf{SEAL} & \textbf{HGCN} & \textbf{TLC-GNN}\\
\midrule
CiteSeer &{\bf 96.32$\pm$0.95} &79.28$\pm$0.80 &79.60$\pm$0.75 &85.05$\pm$0.83 &84.67$\pm$1.80 & 80.79$\pm$0.79 & 94.03$\pm$0.59& 95.11$\pm$0.97 \\
\bottomrule
\end{tabular}
\end{table*}

\begin{table*}[h!]
%\captionsetup{font=large}
\centering
\setlength\tabcolsep{3pt}
\caption{Link prediction in ROC AUC on PPI.\label{ppi_res}}
%\vspace{-2ex}
\begin{tabular}{lccccccc}
\toprule
\textbf{Dataset} & \textbf{BScNets} &\textbf{SkipGNN} & \textbf{DEAL} & \textbf{SAGE} & \textbf{SEAL} &\textbf{GCN} & \textbf{GAT}\\
\midrule
PPI &{\bf 89.19$\pm$0.16} &87.77$\pm$0.17 &88.56$\pm$0.20 &81.15$\pm$0.12 &88.19$\pm$0.19 &83.65$\pm$0.15 &81.79$\pm$0.18 \\
\bottomrule
\end{tabular}
\end{table*}

\begin{table*}[h!]
%\captionsetup{font=footnotesize} %scriptsize
\centering
\captionsetup{justification=centering}
\caption{Hits@20 and Hits@50 score %comparisons 
on OGBs
%OGBL-DDI\\ and OGBL-COLLAB respectively 
with the 5\% and 10\% training sets.\label{ogbl_res}}
%\vspace{-2ex}
\begin{tabular}{lccccc}
\toprule
\textbf{Dataset} & \textbf{Ratio} & \textbf{BScNets} & \textbf{GCN} & \textbf{SAGE} & \textbf{LRGA}\\
\midrule
OGBL-DDI &5\% & {\bf 25.52$\pm$0.03 } &23.30$\pm$0.05  &13.93$\pm$0.05  &22.02$\pm$0.09  \\
OGBL-DDI &10\% & {\bf 31.53$\pm$0.08 } &25.13$\pm$0.06  &26.41$\pm$0.08  &28.69$\pm$0.12  \\
\hline\hline
OGBL-COLLAB &5\% &{\bf 27.09$\pm$0.08} & 24.87$\pm$0.07 &18.26$\pm$0.09   & 21.39$\pm$0.15\\
OGBL-COLLAB &10\% &{\bf 31.00$\pm$0.08} &27.08$\pm$0.05 & 21.71$\pm$0.03 &27.12$\pm$0.10\\
\bottomrule
\end{tabular}
\end{table*}

\begin{table*}[ht!]
%\captionsetup{font=large}
\centering
\caption{Training time (per epoch in seconds). %comparison on Cora and Disease.
\label{training_time}}
%\vspace{-2ex}
\begin{tabular}{lccccc}
\toprule
\textbf{Dataset} & \textbf{BScNets} & \textbf{GCN} & \textbf{SEAL} & \textbf{HGCN} & \textbf{TLC-GNN}\\
\midrule
Cora &$7.04\times 10^{-2}$ &$5.35\times 10^{-2}$ & $5.76\times 10^{-1}$ &$1.13\times 10^{-2}$ & $1.01\times 10^{-1}$ \\
Disease &$3.50 \times 10^{-2}$ &$1.42 \times 10^{-2}$ &$1.97 \times 10^{-1}$ & $1.31 \times 10^{-1}$ &$8.13 \times 10^{-2}$ \\
\bottomrule
\end{tabular}
\end{table*}

\begin{table*}[h!]
%\captionsetup{font=Large}
\centering
\caption{Hyparameters sensitivity analysis of $\delta$ and $\eta$ on Disease dataset.\label{hyparameters_sensitivity}}
%\vspace{-2ex}
\begin{tabular}{lccccc}
\toprule
\textbf{Dataset} & ($\delta = 2, \eta =1$) & ($\delta = 0.1, \eta =1$) & ($\delta = 3, \eta =1$) & ($\delta = 2, \eta = 2$) & ($\delta = 2, \eta = 5$)\\
\midrule
Disease &{\bf 98.60$\pm$0.58} &98.28$\pm$0.50  &98.18$\pm$0.72 &98.31$\pm$0.63 &98.29$\pm$0.63 \\
\bottomrule
\end{tabular}
\end{table*}

\begin{table}[h!]
\caption{Computational costs for generation of incidence matrices (i.e., $\boldsymbol{B}_1$ and $\boldsymbol{B}_2$) and a single training epoch of BScNets.\label{running_time}}
\setlength\tabcolsep{2pt}
\centering
\begin{tabular}{lcc}
\toprule
\multirow{2}{*}{\textbf{Dataset}} & \multicolumn{2}{c}{\textbf{Average Time Taken (sec)}} \\
& Incidence matrices & BScNets (epoch) \\
\midrule
Cora &$7.00\times 10^{-3}$ & $7.04 \times 10^{-2}$\\
PubMed & $4.22 \times 10^{-5}$ & $8.75 \times 10^{-2}$\\
Meetings &$7.55 \times 10^{-4}$  &$5.61 \times 10^{-2}$\\
Phone Calls &$5.08 \times 10^{-5}$ & $4.06 \times 10^{-3}$\\
Airport &$3.36 \times 10^{-4}$ &$2.47 \times 10^{-2}$\\
Disease &$3.81 \times 10^{-6}$ &$3.50 \times 10^{-2}$\\
High School &$1.21 \times 10^{-3}$ & $3.66 \times 10^{-2}$\\
%Teacher Subnetwork &$4.43 \times 10^{-3}$ & $1.07 \times 10^{-2}$\\
Staff Community  &$5.25 \times 10^{-3}$ & $1.14 \times 10^{-2}$\\
\bottomrule
\end{tabular}
\end{table}

% add more results from rebuttal
\subsection{B.5. Additional Experiments}
We now have also run experiments on large datasets and datasets with rich node attributes, i.e., CiteSeer, protein–protein interaction (PPI), OGBL-DDI, and OGBL-COLLAB. CiteSeer is proposed in~\cite{sen2008collective}. PPIs is proposed in~\cite{zitnik2017predicting}. OGBL-DDI and OGBL-COLLAB are proposed in~\cite{hu2020open}. The statistics of these datasets are given in Table~\ref{extra_dataset}. For the experimental results on CiteSeer and PPI, we randomly split edges into 80\%/10\%/10\% for training, validation, and testing, and report the mean average area under the ROC curve (ROC AUC) score (with 5 independent runs). For OGBL-DDI and OGBL-COLLAB, we randomly select $\rho$\% of edges as the training set (where $\rho = \{5, 10\}$), and we use 10\% for both validation and test sets (with an equal \# of true and false edges). The evaluation metric is called Hits@K~\cite{hu2020open} (where K$= \{20, 50\}$). Tables~\ref{citeseer_res} and~\ref{ppi_res} summarize the performance of BScNets and SOAs on CiteSeer and PPIs (we also select several representative baselines to compare including Node2Vec~\cite{grover2016node2vec}, Node2Vec-GCN~\cite{yadati2020nhp}, DEAL~\cite{hao2020inductive}, and SkipGNN~\cite{huang2020skipgnn}). We can observe that BScNets consistently achieves large-margin outperformance over all baselines across both CiteSeer and PPI. The results on OGBL-DDI and OGBL-COLLAB are presented in Table~\ref{ogbl_res}. We observe that BScNets outperforms GCN, SAGE, and LRGA~\cite{puny2020global} on both OGBs.

\begin{table}[ht!]
\centering
\caption{Summary of additional datasets.\label{extra_dataset}}
\begin{tabular}{lccc}
\toprule
\textbf{Dataset} &  \textbf{\# Nodes} & \textbf{\# Edges} & \textbf{\# Features} \\
\midrule
CiteSeer & 3,327 & 4,552 & 3,703 \\
PPI & 1,767 &16,159 &50\\
OGBL-DDI &4,267 &1,334,889 &0\\
OGBL-COLLAB &235,868 &1,285,465 &128\\
\bottomrule
\end{tabular}
\end{table}

Besides transductive learning, following the data splitting setting of~\cite{hao2020inductive}, we also conduct the experiment on Citeseer based on the inductive framework of~\cite{hao2020inductive} with BScNets for the message passing module. Table~\ref{inductive_link_prediction} shows that BScNets also outperforms all baselines for inductive learning.
\begin{table}[ht!]
\centering
\scriptsize
\setlength\tabcolsep{3pt}
\caption{Inductive link prediction in ROC AUC on Citeseer.\label{inductive_link_prediction}}
\begin{tabular}{lccccc}
\toprule
\textbf{Dataset} & \textbf{BScNets} &\textbf{DEAL} & \textbf{SAGE} & \textbf{G2G} & \textbf{MLP}\\
\midrule
Citeseer &{\bf 95.85$\pm$0.17} &93.70$\pm$0.21 & 89.21$\pm$0.26 & 92.13$\pm$0.26 & 87.70$\pm$0.66 \\
\bottomrule
\end{tabular}
\end{table}

\subsection{B.6. Ablation Study on the GCN Layer}
We further perform the ablation study to the effect of the graph convolution operator (i.e., BScNets w/o GCN layer) on Cora and Disease. As Table~\ref{additional_ablation_res} shows, BScNets with GCN layer performs better than w/o GCN layer. The results show importance of capturing essential info at the node-level.
\begin{table}[ht!]
%\captionsetup{font=Large}
\centering
\caption{Ablation study on the GCN layer.\label{additional_ablation_res}}
%\vspace{-2ex}
\begin{tabular}{lcc}
\toprule
\textbf{Dataset} & \textbf{BScNets} & \textbf{BScNets w/o GCN Layer} \\
\midrule
Cora &{\bf 94.90$\pm$0.70} &90.59$\pm$0.75 \\
Disease &{\bf 98.60$\pm$0.58} & 98.13$\pm$0.65 \\
\bottomrule
\end{tabular}
\end{table}

\subsection{B.7. Hyparameters Sensitivity Analysis of $\delta$ and $\eta$}
Table~\ref{hyparameters_sensitivity} shows hyperparameters sensitivity for $\delta$ and $\eta$ on Disease. We find that BScNets with $(\delta =2, \eta = 1)$ yields a relative gain from 0.29\% to 0.43\% in terms of ROC AUC over BScNets with other $(\delta, \eta)$ hyperparameter combinations.

\end{document}